\documentclass[letterpaper]{article} % DO NOT CHANGE THIS
\usepackage{aaai25}  % DO NOT CHANGE THIS
\usepackage{times}  % DO NOT CHANGE THIS
\usepackage{helvet}  % DO NOT CHANGE THIS
\usepackage{courier}  % DO NOT CHANGE THIS
\usepackage[hyphens]{url}  % DO NOT CHANGE THIS
\usepackage{graphicx} % DO NOT CHANGE THIS
\usepackage{natbib}  % DO NOT CHANGE THIS AND DO NOT ADD ANY OPTIONS TO IT
\usepackage{caption} % DO NOT CHANGE THIS AND DO NOT ADD ANY OPTIONS TO IT
\usepackage{algorithm}
\usepackage{algorithmic}

\usepackage{multirow}
\usepackage{newfloat}
\usepackage{listings}
\DeclareCaptionStyle{ruled}{labelfont=normalfont,labelsep=colon,strut=off} % DO NOT CHANGE THIS
\floatstyle{ruled}
\newfloat{listing}{tb}{lst}{}
\floatname{listing}{Listing}
\title{Exploring the Limits of Zero Shot Vision Language Models for Hate Meme Detection: The Vulnerabilities and their Interpretations}
\author {
    Naquee Rizwan, 
    Paramananda Bhaskar, 
    Mithun Das,
    Swadhin Satyaprakash Majhi,\\
    Punyajoy Saha, 
    Animesh Mukherjee
}
\affiliations {
Indian Institute of Technology (IIT), Kharagpur\\
    \{nrizwan@kgpian, pbhaskar@kgpian, animeshm@cse\}.iitkgp.ac.in,\\\{mithundas, punyajoys\}@iitkgp.ac.in, majhiswadhin64@gmail.com
}

\usepackage{xcolor}

\usepackage{bibentry}
\usepackage{colortbl}
\usepackage{xcolor}

\newcommand{\sysI}[1]{\textsc{IDEFICS}}
\newcommand{\sysL}[1]{\textsc{LLaVA-1.5}}
\newcommand{\sysB}[1]{\textsc{InstructBLIP}}
\newcommand{\sysLa}[1]{\textsc{LLaVA}}
\newcommand{\sysGPT}[1]{\textsc{GPT-4o}}

\begin{document}

\maketitle

\begin{abstract}
There is a rapid increase in the use of multimedia content in current social media platforms. One of the highly popular forms of such multimedia content are memes. While memes have been primarily invented to promote funny and buoyant discussions, malevolent users exploit memes to target individuals or vulnerable communities, making it imperative to identify and address such instances of hateful memes. Thus social media platforms are in dire need for active moderation of such harmful content. While manual moderation is extremely difficult due to the scale of such content, automatic moderation is challenged by the need of good quality annotated data to train hate meme detection algorithms. This makes a perfect pretext for exploring the power of modern day vision language models (VLMs) that have exhibited outstanding performance across various tasks. In this paper we study the effectiveness of VLMs in handling intricate tasks such as hate meme detection in a \textit{completely zero-shot setting} so that there is no dependency on annotated data for the task. We perform thorough prompt engineering and query state-of-the-art VLMs using various prompt types to detect hateful/harmful memes. We further interpret the misclassification cases using a novel superpixel based occlusion method. Finally we show that these misclassifications can be neatly arranged into a typology of error classes the knowledge of which should enable the design of better safety guardrails in future\footnote{\textcolor{red}{Accepted at AAAI (ICWSM) 2025}}. Code and other relevant sources are available online\footnote{\url{https://github.com/hate-alert/HateVLMs}}.\\ \textit{\textbf{\textcolor{red}{Warning: Contains potentially offensive content.}}}
\end{abstract}

%%%%%%%%%%%%%%%%%%%%%%%%%%%%%%%%%%%%%%%%%%%%%%%%%%%%%%%%%%%%%%%%%%%%%%%%

\section{Introduction}
\label{section:introduction}
\begin{figure*}[!ht]
    \centering
    \includegraphics[width=1.5\columnwidth]{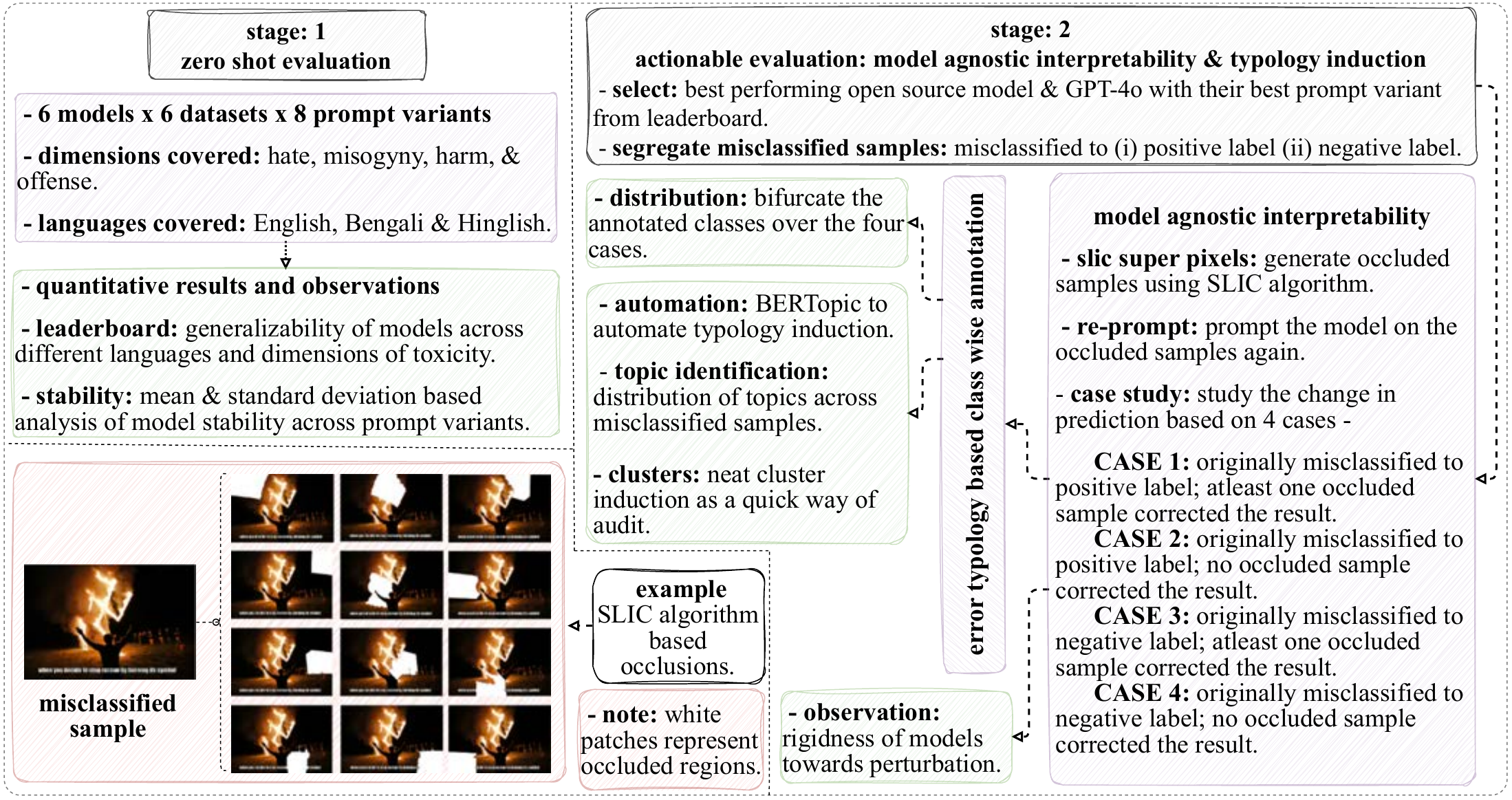}
    \caption{\footnotesize\textbf{\textsc{Pipeline:}} A concise summary and flow of the evaluation and analysis carried out in this work.}
    \label{fig:icwsm_zero_shot}
\end{figure*}

Several \textit{large} vision language models (VLMs) have recently become available to the public~\cite{kheiri2023sentimentgpt,lan2023improving}. A pertinent question is how VLMs perform precisely in the context of hate meme detection~\cite{plaza-del-arco-etal-2023-respectful,van2023detecting} and particularly in a zero-shot setting. The urgency for such systems stem from the exponential growth in multi-modal content on social media platforms and companies like Meta releasing statements of using LLMs/VLMs for content moderation\footnote{\url{https://about.fb.com/news/2025/01/meta-more-speech-fewer-mistakes/}}.  This choice is justified since while manual moderation is nearly impossible, traditional machine learning models can also be not extensively trained for automatic moderation due to the severe lack of labeled hateful memes datasets that are diverse in terms of language, target groups and social setting. This gap in research underscores the need to explore and evaluate the effectiveness of zero-shot VLMs for identifying and mitigating the spread of such content in memes. Note that the zero-shot setting is important here since curating labeled hateful meme datasets that are socially, culturally and target-wise diverse is extremely difficult. \\
In this paper, for the first time, we systematically employ various prompt strategies and input instructions to assess the `power' of well-known VLMs, including \sysI{}~\cite{laurenccon2023obelisc}, \sysL{}~\cite{liu2023visual}, \sysB{}~\cite{dai2023instructblip} and \sysGPT{}~\cite{openaiGPT4} in detecting hateful memes in a fully zero-shot setting. Our work is motivated by the recent literature on zero-shot evaluation of LLMs~\cite{plaza-del-arco-etal-2023-respectful,roy2023probing,saha-etal-2024-zero}. Not only there is an extreme shortage of quality datasets for hateful meme detection but also as we shall show, many memes (due to their implicit nature and nuanced presentation) routinely get incorrectly annotated by humans across benchmark datasets. Thus to make these models usable for moderation in online platforms, it is important to understand their zero-shot capabilities carefully gauging the risks they pose on unseen data so that necessary fixes can be implemented in future. Consequently, our main goal here is to perform interpretability analysis to understand the vulnerabilities faced by these models in a completely zero-shot setup. While fine-tuning the model might improve the results it might also limit the generalizability of the model to the type of data it is being finetuned on. Further, fine-tuning with incorrectly annotated data might result in more harm than good.\\
We evaluate the outputs of these models for four well-known datasets covering hateful, misogynistic, and harmful memes in English and two datasets covering hateful, and offensive memes in Bengali and \textsc{HinGlish}\footnote{Hinglish is a blend of Hindi and English, where speakers mix words, phrases, or sentences from both languages in conversation.} language respectively. The central contributions of this paper are as follows.\\
\noindent\textbf{(i) Systematic evaluation of classification capability of VLMs}: We systematically study the effect of prompt strategies that we use to query these models to understand their strengths and vulnerabilities. In total we investigate as many as \textbf{48} prompts (8 prompt variations across 6 datasets) for each model. This is unlike what is typically done in a majority of studies where the model is queried using one or two prompt variants at most thus limiting the true potential of prompt engineering. Our prompts can be broadly categorized into the following types based on the input and output patterns: input variants can comprise (a) vanilla input, (b) input along with the definition of what is hateful/misogynistic/harmful/offensive, (c) input along with OCR\footnote{\textbf{OCR - Optical Character Recognition}: In memes, this tool is used to extract the embedded textual content. In our work, we use the OCR extracted text already present in the datasets.} text, (d) input along with definition and OCR text; output variants can be (a) vanilla output, (b) output along with an explanation. We observe that prompt variants that are most successful in eliciting correct responses vary across models and datasets; nevertheless, in many of the cases \textit{OCR text alone} or \textit{OCR text with definition} works well.\\
\noindent\textbf{(ii) Interpretation of misclassified results}: In order to understand the reasons for the misclassifications done by a model we present a \textbf{novel} superpixel based occlusion strategy to occlude different parts of an originally mispredicted meme. We note if these occlusions result in a change in the model prediction. If they indeed do, then one can conclude that the occluded parts play an important role in the decision making process of the model. This approach allows us to interpret the failure cases of the model and pinpoints to the regions of the memes that plays a key role in confusing the model predictions. Interestingly, we also find evidences of cases where the ground-truth annotations might themselves have been wrong, per our judgement, as opposed to the model predictions.\\
\noindent\textbf{(iii) Typology of misclassifications}: The final question that we ask in the paper is whether one can systematically organise the misclassifications of the model so that constructive suggestions could be brewed from them to re-engineer the safety guardrails of the VLMs. To this purpose, we cluster the misclassified memes using multi-modal topic modeling thereby inducing a \textit{typology} of error patterns. Interestingly, this typology seems to highly align with the different kinds of superpixel based interpretations that we obtain. This typology can be thought of as the `Achilles heel' of a VLM against which it needs to be safeguarded in future.\\ 
Overall, our study has a far larger scope than the standard objective of identifying the best all-purpose VLMs. It strives to rather choose the best prompt variant across different models using a thorough and principled prompt engineering approach. Further it lays a foundation to identify interpretable typological categories of hateful memes that the VLMs are most vulnerable to. These induced topics can be used for \textit{actionable evaluation}~\cite{vilar-etal-2006-error} to improve the performance of VLMs by implementing safety guardrails without fine-tuning the models repeatedly which typically comes with a huge compute cost. We outline the summary and flow of the paper in Figure~\ref{fig:icwsm_zero_shot}.

%%%%%%%%%%%%%%%%%%%%%%%%%%%%%%%%%%%%%%%%%%%%%%%%%%%%%%%%%%%%%%%%%%%%%%%%

\section{Related works}
\noindent\textbf{Hate meme detection}: A growing body of research in recent years focused on hate meme detection~\cite{gomez2020exploring,kirk2021memes,shang2021aomd}. Several datasets and models have been developed, encompassing various dimensions, including hateful content detection~\cite{kiela2020hateful}, misogyny detection~\cite{fersini2022semeval}, cyberbullying detection~\cite{maity2022multitask}, harmful meme detection~\cite{pramanick2021detecting,pramanick2021momenta}, and many more~\cite{chandra2021subverting, lin2024goatbench} including other languages~\cite{das2023banglaabusememe,hossain2024deciphering,kumari-etal-2024-cm}.\\
\noindent\textbf{Vision language models}:
\sysI{}~\cite{laurenccon2023obelisc}, \sysL{}~\cite{liu2023visual}, \sysB{}~\cite{dai2023instructblip}, Flamingo~\cite{alayrac2022flamingo}, and \sysGPT{}~\cite{openaiGPT4} are popular vision language models widely used for tasks like sentiment analysis~\cite{kheiri2023sentimentgpt}, question answering~\cite{lan2023improving}, and hate meme detection~\cite{van2023detecting}. However, exploring hate meme detection using VLMs is limited~\cite{van2023detecting,  lin2024goatbench}, particularly in the context of different prompt scenarios, different model setups \& languages and thorough interpretation of results.\\
\noindent\textbf{Interpretability:} With the advent of deep learning, several efforts are made to explain the working of the model's predictions~\cite{8237336, Gyevnar2023BridgingTT, 10.1145/2939672.2939778, inproceedings}. Recently, there have also been efforts to evaluate vision-language multimodal systems~\cite{inbook,zhao-etal-2022-explainable,lei2023lico}. However, memes are a special type of image where the actual context is inherently decided by the combination of image and the embedded text within the meme. There are a handful of works present on understanding memes as well~\cite{10.1145/3462244.3479949, 10.1145/3485447.3512260, Hee2023DecodingTU}. However none of these works present the study of important regions in the meme that lead to model's prediction with respect to an input prompt given to the VLMs. More importantly, to the best of our knowledge, there are no works presently that can interpret the predictions of closed source VLMs like \sysGPT{}. In this work, we propose a \textbf{\textit{novel} model agnostic superpixel} based approach to effectively understand the important regions in the meme leading to model's prediction corresponding to an input prompt.

%%%%%%%%%%%%%%%%%%%%%%%%%%%%%%%%%%%%%%%%%%%%%%%%%%%%%%%%%%%%%%%%%%%%%%%%

\section{Datasets and metrics}
\label{section:datsets_and_metrics}

\textbf{{Datasets}:} This section introduces the six datasets we have utilized to explore the performance of Vision-Language Models (VLMs). These datasets cover four types of memes: hate, misogyny, harmful, or offensive content spanning three languages: English, Bengali \& \textsc{HinGlish}. (see Table \ref{tab:dataDist} for details).

\begin{table}[h]
\centering
\scriptsize
\begin{tabular}{c|cc|c}
% \hline
\textbf{Dataset}        & \multicolumn{2}{c|}{\textbf{Label distribution}} & \textbf{Total}        \\ \hline
\multirow{2}{*}{FHM}    & \multicolumn{1}{c}{Hateful}          & 250      & \multirow{2}{*}{500}  \\ 
                        & \multicolumn{1}{c}{Not hateful}      & 250      &                       \\ \hline
\multirow{2}{*}{MAMI}   & \multicolumn{1}{c}{Misogynous}           & 500      & \multirow{2}{*}{1000} \\
                        & \multicolumn{1}{c}{Not misogynous}       & 500      &                       \\ \hline
\multirow{2}{*}{HARM-P} & \multicolumn{1}{c}{Harmful}          & 173      & \multirow{2}{*}{355}  \\
                        & \multicolumn{1}{c}{Not harmful}      & 182      &                       \\ \hline
\multirow{2}{*}{HARM-C} & \multicolumn{1}{c}{Harmful}          & 124      & \multirow{2}{*}{354}  \\
                        & \multicolumn{1}{c}{Not harmful}      & 230      &                       \\ \hline
\multirow{2}{*}{BHM} & \multicolumn{1}{c}{Hateful}          & 266      & \multirow{2}{*}{711}  \\
                        & \multicolumn{1}{c}{Not hateful}      & 445      &                       \\ \hline
\multirow{2}{*}{\textsc{HinGlish}} & \multicolumn{1}{c}{Offensive}          & 250      & \multirow{2}{*}{500}  \\
                        & \multicolumn{1}{c}{Not offensive}      & 250      &                       \\
\end{tabular}
\caption{\textbf{\textsc{Datasets:}} Label distribution for each dataset.}
\label{tab:dataDist}
\end{table}

\noindent\textbf{(1) Facebook hateful memes (FHM)}: The FHM dataset introduced by Facebook AI~\cite{kiela2020hateful} is a collection of memes designed to help researchers develop tools for identifying and removing hateful content online. The dataset contains more than 10K memes labeled \textit{hateful} and \textit{not-hateful}, covering various targets, including race, ethnicity, religion, gender, sexual orientation, and disability. We use a random sample of 500 memes\footnote{Note that for this dataset the test set was removed by the authors after the competition; therefore we used the \textit{dev} split.} in order to test the VLMs in the zero-shot setting. 

\noindent\textbf{(2) Multimedia automatic misogyny identification (MAMI):} The MAMI~\cite{fersini2022semeval} dataset was created and shared as part of SemEval-2022 Task 5. Unlike the FHM dataset, the MAMI dataset focuses on identifying misogyny in online memes. The dataset contains 11K memes, of which 1K memes are in the test set, and we conduct all our experiments considering only the test set. Each meme has a binary label -- \textit{misogynous} or \textit{not misogynous} -- which we use for our experiments.

\noindent\textbf{(3) Harmful memes}: `Harmful' is a more general term compared to `offensive' and `hateful'. While an offensive or hateful meme is harmful, not all harmful memes are necessarily hateful or offensive. We utilize the \textbf{HARM-P}~\cite{pramanick2021momenta}  (related to US politics) and \textbf{HARM-C}~\cite{pramanick2021detecting} (related to COVID-19) datasets for our experiments. Both datasets contain more than 3.5K memes. For our study, we only consider the test sets. The original labels of both datasets have three classification labels: \textit{not harmful}, \textit{somewhat harmful}, and \textit{very harmful}. To maintain consistency with our binary classification experiments, we have merged \textit{somewhat harmful} and \textit{very harmful} into a single category labeled as \textit{harmful}.

\noindent\textbf{(4) Bangla hateful memes}: The BHM dataset was proposed 
 in a recent work by~\cite{hossain2024deciphering} and contains over 7K memes in Bengali language. These memes are labelled as \textit{hateful} or \textit{not-hateful} and are specifically prevalent among Bengali speakers in India \& Bangladesh, covering a variety of discussion topics. We use the test split with 711 memes for our experiments.

\noindent\textbf{(5) \textsc{HinGlish} offensive memes}: \textsc{HinGlish} is a language commonly spoken within Indian subcontinent and is a mix of Hindi with English. A recent work by \cite{kumari-etal-2024-cm} introduced a dataset in \textsc{HinGlish} which contains \textit{offensive} or \textit{not-offensive} memes. Although the paper mentions around 7k samples, just over 2k memes have been made public\footnote{\url{https://github.com/Gitanjali1801/CM_MEMES}}. We reached out to the authors of the paper for the complete dataset but without any success. The memes surround the activities which are prevalent within Indian subcontinent. Since the full dataset has not been made public, we randomly sample 250 memes from each of the offensive and not offensive classes.

\noindent\textbf{Metrics:} As we perform binary classification tasks, we measure the models' performance using \textbf{accuracy}, and  \textbf{macro-F1 score} as metrics.

%%%%%%%%%%%%%%%%%%%%%%%%%%%%%%%%%%%%%%%%%%%%%%%%%%%%%%%%%%%%%%%%%%%%%%%%

\section{Models}
\label{section:models}
We ran our experiments on a total of \textbf{six} different models.\\
\noindent\textbf{\sysI{}}: \sysI{}~\cite{laurenccon2023obelisc} which closely follows the architecture of Flamingo, is trained on open source datasets like \textsc{OBELICS} and \textsc{LAION}. It combines two frozen uni-modal backbones which are, \textsc{LLaMA} as the language model and \textsc{OpenClip} as the vision encoder. 
We used instruction fine-tuned \sysI{} 9B model with the checkpoint \textit{\textbf{HuggingFaceM4/idefics-9b-instruct}} 
for our experiments.\\
\noindent\textbf{\sysL{}}: \sysL{}~\cite{liu2023visual} is an enhanced version of \sysLa{}. \sysLa{} combines \textsc{LLaMA}/Vicuna as the language model and CLIP as the vision encoder. Compared to \sysLa{}, \sysL{} has enhanced capabilities due to the addition of an MLP vision-language connector and integration of academic task-oriented data. We have used two different \sysL{} models with 7B and 13B parameters. The checkpoints of these models are \textit{\textbf{llava-hf/llava-1.5-7b-hf}} and \textit{\textbf{llava-hf/llava-1.5-13b-hf}}.\\
\noindent\textbf{\sysB{}}: \sysB{}~\cite{dai2023instructblip} is an instruction fine-tuned model that uses the same architecture as BLIP-2 with a small but significant difference. It uses frozen Flan-T5/Vicuna as the language model and a vision transformer as the image encoder. Extending BLIP-2, \sysB{} proposes an instruction-aware Q-Former module. As additional inputs, the model takes instruction text tokens which interacts with the query embeddings via the self-attention layer of the Q-Former.
We have used two different \sysB{} models with Vicuna 7B  and Flan-T5-xl  as backbone language models. The checkpoints of these models are \textit{\textbf{Salesforce/instructblip-vicuna-7b}} and \textit{\textbf{Salesforce/instructblip-flan-t5-xl}} respectively.\\
\noindent\textbf{\sysGPT{}}: \sysGPT{}~\cite{openaiGPT4} \texttt{(`o'- omni)} is a model released by OpenAI that takes input an arbitrary combination of image or text and generates textual output. Compared to previous OpenAI models, \sysGPT{} is comparatively better at vision understanding. In our work, this is the only closed source model which we have considered apart from the five open source models. We would like to thank \textit{Microsoft Azure} for providing us with the credits to use their OpenAI services. We used the provided APIs to infer \sysGPT{} from their documentation to run our experiments\footnote{https://learn.microsoft.com/en-us/azure/ai-services/openai/}
and for our case, we used a combination of single image with text as input to remain consistent with the \textit{zero-shot} setup.

\noindent\textbf{\textsc{Baseline:}} Our work focuses on the evaluation of zero-shot capability of VLMs. To have a fair comparison with a very recent prior work presented in~\cite{cao2023promptingmultimodalhatefulmeme}, we consider two different baselines: \textbf{(i)} complete zero-shot assessment, and \textbf{(ii)} fine-tuned only on FHM (a very diverse dataset) for two epochs (instead of ten; to be very near to the zero-shot setup) and evaluation across all six datasets. These setups not only allow for fair evaluation, but also help in assessing the generalization capability, which is often ignored. Two samples (one negative and one positive) are also provided with the test sample in the prompt, as noted in the mentioned paper; hence we present results as mean and standard deviation across three different seeds for each setup.

%%%%%%%%%%%%%%%%%%%%%%%%%%%%%%%%%%%%%%%%%%%%%%%%%%%%%%%%%%%%%%%%%%%%%%%%

\section{Prompts}

This section presents the array of prompt variants employed in our work. A concise summary of representative examples for the prompt variants is provided in Appendix\ref{sec:appendixA}, while detailed information for each variant is discussed below.\\
\noindent\textbf{Input patterns}: We run our experiments on four different input patterns, which are as follows.\\
\noindent\textit{\textbf{Vanilla input}}: Following~\cite{roy2023probing}, we use a prompt template to instruct the model to classify the given meme into a label from a predefined \texttt{list\_of\_labels}. However, in our scenario, the \texttt{list\_of\_labels} is only restricted to binary labels. In addition, we supply two \texttt{example\_outputs} (one label per line for positive and negative samples) to assist the models in generating appropriate answers. In our case, `positive' denotes content deemed hateful, misogynistic, harmful, or offensive based on the dataset passed to the model.\\
\noindent(\textbf{+}) \textit{\textbf{Definition input}}: For vanilla prompts, we assumed that VLMs are to some extent aware of the labels for classifying the input image. Here, we take a step further and add the \texttt{definition} of labels as an additional context to the VLMs. Our intuition was similar to~\cite{roy2023probing}, i.e., the \texttt{definition} can help the VLMs understand the classification tasks better. We picked and added one line of \texttt{definition} from the corresponding dataset for all \texttt{list\_of\_labels} (positive and negative in our case). We provide definitions of the labels for each dataset in Appendix\ref{sec:dataset_definitions}.\\
\noindent(\textbf{+}) \textit{\textbf{OCR input}}: In a meme, multi-modality, i.e., embedded text and image play very crucial role in classification, similar to the works~\cite{pramanick2021detecting,das2023banglaabusememe}. We therefore add \texttt{ocr\_extracted\_text} in the vanilla prompt. Our intuition was that the models would further be better in understanding the contexts with this addition and would be more successful in classifying the input meme as per the \texttt{list\_of\_labels}. We provide the \texttt{ocr\_extracted\_text} enclosed within three back-ticks for the model to distinguish it from other texts in the prompt.\\
\noindent(\textbf{+}) \textit{\textbf{Definition \& OCR input}}: Here, we combine both \texttt{definition} and \texttt{ocr\_extracted\_text} with vanilla prompt and pass it as input prompt for our experiment. We use all intuitions discussed above in previous prompt variants and assume that this prompt would provide the models with deeper contexts for the classification task. Moreover, in this setup the order of the prompt text is the \texttt{definition} followed by the \texttt{ocr\_extracted\_text}.\\
\noindent\textbf{Output patterns}: We run our experiment on two different output patterns which are noted below.\\
\noindent\textit{\textbf{Vanilla output}}: In this case, we prompt the model to generate as output only the correct class label from the \texttt{list\_of\_labels} corresponding to different datasets as mentioned in Table~\ref{tab:dataDist}.\\ 
\noindent(\textbf{+}) \textit{\textbf{Explanation output}}: Adding to the above case of vanilla output, we prompt the model to further explain the rationale (within 30 words) based on which it made a prediction.\\
Thus we run a total of \textit{eight} prompts for each dataset and for each model setup by running four input patterns $\times$ two output patterns.\\
The definition of hate speech, the explanation as to why a post is hateful, and the OCR text (often indicating the victim community) are a hate meme's three most important components. We believe that prompts should at least include information about these components, as has also been noted in~\cite{roy2023probing}. We agree that there can be other variants but at meta level the above three units should always be considered while framing a prompt for hate meme detection.

%%%%%%%%%%%%%%%%%%%%%%%%%%%%%%%%%%%%%%%%%%%%%%%%%%%%%%%%%%%%%%%%%%%%%%%%

\section{Experimental setup}

For all the models, we use a batch size of 1. We manually tune the temperature values and set them to 1.0 for the \sysI{}, \sysL{} 7B \& 13B, \sysGPT{} models, and 0.8 for the \sysB{} models. The temperature parameter controls how random the generated output would be. However, with lower temperatures, we observed inferior performance of these models. As noted earlier, we experiment with eight different prompts on six datasets, studying them across six models. In short, we run \textbf{48} prompts per model and \textbf{288} prompts across all six models. All the open source models are coded in Python using the PyTorch library while we have no information about \sysGPT{} since it is a  proprietary model. For open source models, we utilize 2xT4 GPUs from Kaggle, providing a total of 15GB memory on each GPU with a usage limit of 30hrs/week We provide further setup details for these models in Appendix\ref{sec:reproducibility}. We present the detailed results in the following section.

%%%%%%%%%%%%%%%%%%%%%%%%%%%%%%%%%%%%%%%%%%%%%%%%%%%%%%%%%%%%%%%%%%%%%%%%

\section{Results}

\begin{table*}[!hbt]
\centering
\footnotesize
\setlength{\tabcolsep}{4.75pt}
\begin{tabular}{cccccccccccccc}
\hline
\multicolumn{2}{|c|}{\textbf{Strategies}}                                   & \multicolumn{2}{c|}{\textbf{FHM}}                                          & \multicolumn{2}{c|}{\textbf{MAMI}}                   & \multicolumn{2}{c|}{\textbf{HARM-C}}                 & \multicolumn{2}{c|}{\textbf{HARM-P}}                 & \multicolumn{2}{c|}{\textbf{BHM}}                                 & \multicolumn{2}{c|}{\textsc{\textbf{HinGlish}}}                            \\ \hline
\multicolumn{1}{|c}{\textbf{in}}        & \multicolumn{1}{c|}{\textbf{out}} & {\textbf{acc}}                         & \multicolumn{1}{c|}{\textbf{mf1}}   & {\textbf{acc}}   & \multicolumn{1}{c|}{\textbf{mf1}}   & {\textbf{acc}}   & \multicolumn{1}{c|}{\textbf{mf1}}   & {\textbf{acc}}   & \multicolumn{1}{c|}{\textbf{mf1}}   & {\textbf{acc}}                  & \multicolumn{1}{c|}{\textbf{mf1}} & {\textbf{acc}}                  & \multicolumn{1}{c|}{\textbf{mf1}} \\ \hline
                                        &                                   & \multicolumn{12}{c}{\scriptsize\textbf{\sysI{} 9B}}                                                                                                                                                                                                                                                                                                                                                \\ \hline
\multicolumn{1}{|c}{\textbf{vn}}        & \multicolumn{1}{c|}{\textbf{vn}}  & {53.2}          & \multicolumn{1}{c|}{48.84}          & 50.5           & \multicolumn{1}{c|}{34.96}          & 62.99          & \multicolumn{1}{c|}{53.64}          & 50.42          & \multicolumn{1}{c|}{49.68}          & 37.13                        & \multicolumn{1}{c|}{27.25}       & \underline{54.8}                         & \multicolumn{1}{c|}{\underline{50.46}}       \\
\multicolumn{1}{|c}{\textbf{def}}       & \multicolumn{1}{c|}{\textbf{vn}}  & {\colorbox[HTML]{DCDCDC}{50.14}}         & \multicolumn{1}{c|}{\colorbox[HTML]{DCDCDC}{33.4}}           & 50             & \multicolumn{1}{c|}{33.33}          & \colorbox[HTML]{DCDCDC}{44.49}          & \multicolumn{1}{c|}{\colorbox[HTML]{DCDCDC}{43.32}}          & \colorbox[HTML]{DCDCDC}{51.12}          & \multicolumn{1}{c|}{\colorbox[HTML]{DCDCDC}{50.34}}          & 37.38                        & \multicolumn{1}{c|}{27.21}       & 51.2                         & \multicolumn{1}{c|}{41.25}       \\
\multicolumn{1}{|c}{\textbf{ocr}}       & \multicolumn{1}{c|}{\textbf{vn}}  & {\underline{58}}            & \multicolumn{1}{c|}{\underline{57.64}}          & \underline{53.2}           & \multicolumn{1}{c|}{42.58}          & 64.31          & \multicolumn{1}{c|}{\textbf{61.64}} & \textbf{63.38} & \multicolumn{1}{c|}{\textbf{63.1}}  & \underline{61.32}                        & \multicolumn{1}{c|}{49.14}       & 50.2                         & \multicolumn{1}{c|}{48.92}       \\
\multicolumn{1}{|c}{\textbf{def + ocr}} & \multicolumn{1}{c|}{\textbf{vn}}  & 52.02                                & \multicolumn{1}{c|}{41.29}          & 50.1           & \multicolumn{1}{c|}{33.56}          & 45.35          & \multicolumn{1}{c|}{45.29}          & 53.67          & \multicolumn{1}{c|}{53.55}          & 37.41                        & \multicolumn{1}{c|}{27.23}       & 51.1                         & \multicolumn{1}{c|}{48.05}       \\
\multicolumn{1}{|c}{\textbf{vn}}        & \multicolumn{1}{c|}{\textbf{ex}}  & {51.2}          & \multicolumn{1}{c|}{43.16}          & 50.1           & \multicolumn{1}{c|}{33.56}          & 51.13          & \multicolumn{1}{c|}{50.01}          & 47.61          & \multicolumn{1}{c|}{46.97}          & 37.41                        & \multicolumn{1}{c|}{27.23}       & 50.8                         & \multicolumn{1}{c|}{47.46}       \\
\multicolumn{1}{|c}{\textbf{def}}       & \multicolumn{1}{c|}{\textbf{ex}}  & {50.6}          & \multicolumn{1}{c|}{34.65}          & 50.9           & \multicolumn{1}{c|}{38.91}          & 35.04          & \multicolumn{1}{c|}{28.66}          & 50.14          & \multicolumn{1}{c|}{46.55}          & {37.5}  & \multicolumn{1}{c|}{27.27}       & {49}   & \multicolumn{1}{c|}{36.62}       \\
\multicolumn{1}{|c}{\textbf{ocr}}       & \multicolumn{1}{c|}{\textbf{ex}}  & 57.6                                 & \multicolumn{1}{c|}{57.45}          & 50.15          & \multicolumn{1}{c|}{\underline{50.13}}          & \underline{64.41}          & \multicolumn{1}{c|}{39.92}          & 48.17          & \multicolumn{1}{c|}{48.17}          & 52.2                         & \multicolumn{1}{c|}{\underline{50.67}}       & 51.8                         & \multicolumn{1}{c|}{41.6}        \\
\multicolumn{1}{|c}{\textbf{def + ocr}} & \multicolumn{1}{c|}{\textbf{ex}}  & {49.8}          & \multicolumn{1}{c|}{38.15}          & 49.4           & \multicolumn{1}{c|}{36.69}          & 51.84          & \multicolumn{1}{c|}{43}             & 53.39          & \multicolumn{1}{c|}{47.22}          & {38.69} & \multicolumn{1}{c|}{30.71}       & {48.8}  & \multicolumn{1}{c|}{36.23}       \\ \hline
                                        &                                   & \multicolumn{12}{c}{\scriptsize\textbf{\sysL{} 13B}}                                                                                                                                                                                                                                                                                                                                                 \\ \hline
\multicolumn{1}{|c}{\textbf{vn}}        & \multicolumn{1}{c|}{\textbf{vn}}  & 55.95                                & \multicolumn{1}{c|}{52.27}          & 62.3           & \multicolumn{1}{c|}{58.09}          & 53.95          & \multicolumn{1}{c|}{53.76}          & 54.93          & \multicolumn{1}{c|}{54.32}          & \colorbox[HTML]{DCDCDC}{43.67}                        & \multicolumn{1}{c|}{\colorbox[HTML]{DCDCDC}{43.17}}       & \colorbox[HTML]{DCDCDC}{51.67}                        & \multicolumn{1}{c|}{\colorbox[HTML]{DCDCDC}{49.92}}       \\
\multicolumn{1}{|c}{\textbf{def}}       & \multicolumn{1}{c|}{\textbf{vn}}  & 57.96                                & \multicolumn{1}{c|}{57.46}          & 60.84          & \multicolumn{1}{c|}{60.63}          & 54.76          & \multicolumn{1}{c|}{54.53}          & 54.79          & \multicolumn{1}{c|}{53.95}          & 59.94                        & \multicolumn{1}{c|}{\textbf{57.47}}       & 54.33                        & \multicolumn{1}{c|}{53.35}       \\
\multicolumn{1}{|c}{\textbf{ocr}}       & \multicolumn{1}{c|}{\textbf{vn}}  & 54.8                                 & \multicolumn{1}{c|}{52.59}          & 55.22          & \multicolumn{1}{c|}{51.38}          & \underline{61.61}          & \multicolumn{1}{c|}{56.88}          & \underline{59.57}          & \multicolumn{1}{c|}{\underline{58.62}}          & \colorbox[HTML]{DCDCDC}{61.6}                         & \multicolumn{1}{c|}{\colorbox[HTML]{DCDCDC}{54.41}}       & \underline{56.22}                        & \multicolumn{1}{c|}{53}         \\
\multicolumn{1}{|c}{\textbf{def + ocr}} & \multicolumn{1}{c|}{\textbf{vn}}  & \underline{58.57}                                & \multicolumn{1}{c|}{\underline{58.33}}          & \textbf{67.56} & \multicolumn{1}{c|}{\textbf{67.55}} & 58.63          & \multicolumn{1}{c|}{\underline{58.07}}          & 56.12          & \multicolumn{1}{c|}{55.61}          & \textbf{64.25}                        & \multicolumn{1}{c|}{50.7}        & 55.88                        & \multicolumn{1}{c|}{\underline{53.69}}       \\
\multicolumn{1}{|c}{\textbf{vn}}        & \multicolumn{1}{c|}{\textbf{ex}}  & {56.61}         & \multicolumn{1}{c|}{55.89}          & 61.92          & \multicolumn{1}{c|}{61.9}           & 55.81          & \multicolumn{1}{c|}{45.09}          & 54.31          & \multicolumn{1}{c|}{49.73}          & {46.33} & \multicolumn{1}{c|}{46.33}       & {\colorbox[HTML]{DCDCDC}{54.04}} & \multicolumn{1}{c|}{\colorbox[HTML]{DCDCDC}{52.13}}       \\
\multicolumn{1}{|c}{\textbf{def}}       & \multicolumn{1}{c|}{\textbf{ex}}  & {50.51}         & \multicolumn{1}{c|}{36.89}          & 62.59          & \multicolumn{1}{c|}{62.58}          & 42.86          & \multicolumn{1}{c|}{40.59}          & 50.28          & \multicolumn{1}{c|}{42.57}          & {36.59} & \multicolumn{1}{c|}{27.85}       & {51.68} & \multicolumn{1}{c|}{42.97}       \\
\multicolumn{1}{|c}{\textbf{ocr}}       & \multicolumn{1}{c|}{\textbf{ex}}  & 57.5                                 & \multicolumn{1}{c|}{57.5}           & 64.16          & \multicolumn{1}{c|}{63.97}          & 54.05          & \multicolumn{1}{c|}{51.65}          & 58             & \multicolumn{1}{c|}{56.09}          & 52.15                        & \multicolumn{1}{c|}{51.47}       & 50.91                        & \multicolumn{1}{c|}{36.79}       \\
\multicolumn{1}{|c}{\textbf{def + ocr}} & \multicolumn{1}{c|}{\textbf{ex}}  & {49.7}          & \multicolumn{1}{c|}{36.46}          & 63.03          & \multicolumn{1}{c|}{62.21}          & 43.55          & \multicolumn{1}{c|}{40.74}          & 50             & \multicolumn{1}{c|}{41.23}          & {38.29} & \multicolumn{1}{c|}{29.78}       & {53.31} & \multicolumn{1}{c|}{52.19}       \\ \hline
                                        &                                   & \multicolumn{12}{c}{\scriptsize\textbf{\sysL{} 7B}}                                                                                                                                                                                                                                                                                                                                                  \\ \hline
\multicolumn{1}{|c}{\textbf{vn}}        & \multicolumn{1}{c|}{\textbf{vn}}  & 50                                   & \multicolumn{1}{c|}{33.33}          & 50.8           & \multicolumn{1}{c|}{35.25}          & 64.97          & \multicolumn{1}{c|}{39.38}          & 51.27          & \multicolumn{1}{c|}{33.89}          & \colorbox[HTML]{DCDCDC}{57.69}                        & \multicolumn{1}{c|}{\colorbox[HTML]{DCDCDC}{38.77}}       & 55.29                        & \multicolumn{1}{c|}{53.84}       \\
\multicolumn{1}{|c}{\textbf{def}}       & \multicolumn{1}{c|}{\textbf{vn}}  & 52.8                                 & \multicolumn{1}{c|}{46.79}          & 50.82          & \multicolumn{1}{c|}{41.81}          & \textbf{67.35} & \multicolumn{1}{c|}{58.12}          & 52.46          & \multicolumn{1}{c|}{43.25}          & \colorbox[HTML]{DCDCDC}{54.29}                        & \multicolumn{1}{c|}{\colorbox[HTML]{DCDCDC}{54.05}}       & 50.77                        & \multicolumn{1}{c|}{34.46}       \\
\multicolumn{1}{|c}{\textbf{ocr}}       & \multicolumn{1}{c|}{\textbf{vn}}  & 53.31                                & \multicolumn{1}{c|}{46.32}          & 53.4           & \multicolumn{1}{c|}{41.17}          & 65.25          & \multicolumn{1}{c|}{40.25}          & 51.27          & \multicolumn{1}{c|}{33.89}          & \colorbox[HTML]{DCDCDC}{60.88}                        & \multicolumn{1}{c|}{\colorbox[HTML]{DCDCDC}{46.55}}       & \textbf{57.47}                        & \multicolumn{1}{c|}{\textbf{54.63}}       \\
\multicolumn{1}{|c}{\textbf{def + ocr}} & \multicolumn{1}{c|}{\textbf{vn}}  & 55.6                                 & \multicolumn{1}{c|}{50.39}          & 62.7           & \multicolumn{1}{c|}{60.44}          & 65.25          & \multicolumn{1}{c|}{\underline{59.93}}          & \underline{54.93}          & \multicolumn{1}{c|}{52.7}           & \colorbox[HTML]{DCDCDC}{57.27}                        & \multicolumn{1}{c|}{\colorbox[HTML]{DCDCDC}{50.64}}       & 51.02                        & \multicolumn{1}{c|}{35.22}       \\
\multicolumn{1}{|c}{\textbf{vn}}        & \multicolumn{1}{c|}{\textbf{ex}}  & {50.4}          & \multicolumn{1}{c|}{36.18}          & 55.1           & \multicolumn{1}{c|}{48.37}          & 64.97          & \multicolumn{1}{c|}{39.38}          & 51.55          & \multicolumn{1}{c|}{34.53}          & {\colorbox[HTML]{DCDCDC}{53.2}}  & \multicolumn{1}{c|}{\colorbox[HTML]{DCDCDC}{49.47}}       & {52.29} & \multicolumn{1}{c|}{41.39}       \\
\multicolumn{1}{|c}{\textbf{def}}       & \multicolumn{1}{c|}{\textbf{ex}}  & {55}            & \multicolumn{1}{c|}{53.91}          & 54.7           & \multicolumn{1}{c|}{46.12}          & 48.02          & \multicolumn{1}{c|}{47.28}          & 49.86          & \multicolumn{1}{c|}{47.52}          & {\colorbox[HTML]{DCDCDC}{39.01}} & \multicolumn{1}{c|}{\colorbox[HTML]{DCDCDC}{29.23}}       & {50}    & \multicolumn{1}{c|}{33.33}       \\
\multicolumn{1}{|c}{\textbf{ocr}}       & \multicolumn{1}{c|}{\textbf{ex}}  & 51.2                                 & \multicolumn{1}{c|}{41.45}          & 52.7           & \multicolumn{1}{c|}{40.89}          & 64.97          & \multicolumn{1}{c|}{39.38}          & 51.55          & \multicolumn{1}{c|}{35.03}          & \underline{52.81}                        & \multicolumn{1}{c|}{\underline{51.3}}        & 51.1                         & \multicolumn{1}{c|}{35.57}       \\
\multicolumn{1}{|c}{\textbf{def + ocr}} & \multicolumn{1}{c|}{\textbf{ex}}  & {\textbf{60}}   & \multicolumn{1}{c|}{\textbf{59.98}} & \underline{63.6}           & \multicolumn{1}{c|}{\underline{63.48}}          & 60.45          & \multicolumn{1}{c|}{59.03}          & 54.08          & \multicolumn{1}{c|}{\underline{54.07}}          & {46.14} & \multicolumn{1}{c|}{45.19}       & 50.2                         & \multicolumn{1}{c|}{33.78}       \\ \hline
                                        &                                   & \multicolumn{12}{c}{\scriptsize\textbf{\sysB{} \textsc{Vicuna 7B}}}                                                                                                                                                                                                                                                                                                                                    \\ \hline
\multicolumn{1}{|c}{\textbf{vn}}        & \multicolumn{1}{c|}{\textbf{vn}}  & \colorbox[HTML]{DCDCDC}{54.14}                                & \multicolumn{1}{c|}{\colorbox[HTML]{DCDCDC}{38.59}}          & \colorbox[HTML]{DCDCDC}{46.86}          & \multicolumn{1}{c|}{\colorbox[HTML]{DCDCDC}{31.91}}          & \colorbox[HTML]{DCDCDC}{44.25}          & \multicolumn{1}{c|}{\colorbox[HTML]{DCDCDC}{40.16}}          & \colorbox[HTML]{DCDCDC}{43.98}          & \multicolumn{1}{c|}{\colorbox[HTML]{DCDCDC}{33.42}}          & \colorbox[HTML]{DCDCDC}{60.94}                        & \multicolumn{1}{c|}{\colorbox[HTML]{DCDCDC}{39.23}}       & \colorbox[HTML]{DCDCDC}{51.76}                        & \multicolumn{1}{c|}{\colorbox[HTML]{DCDCDC}{50.03}}       \\
\multicolumn{1}{|c}{\textbf{def}}       & \multicolumn{1}{c|}{\textbf{vn}}  & 51.12                                & \multicolumn{1}{c|}{34.55}          & 49.74          & \multicolumn{1}{c|}{34.44}          & \colorbox[HTML]{DCDCDC}{49.12}          & \multicolumn{1}{c|}{\colorbox[HTML]{DCDCDC}{48.65}}          & \underline{48.63}          & \multicolumn{1}{c|}{41.05}          & \colorbox[HTML]{DCDCDC}{61.49}                        & \multicolumn{1}{c|}{\colorbox[HTML]{DCDCDC}{42.44}}       & \colorbox[HTML]{DCDCDC}{50.12}                        & \multicolumn{1}{c|}{\colorbox[HTML]{DCDCDC}{41.04}}       \\
\multicolumn{1}{|c}{\textbf{ocr}}       & \multicolumn{1}{c|}{\textbf{vn}}  & 50.1                                 & \multicolumn{1}{c|}{33.73}          & 48.37          & \multicolumn{1}{c|}{33.94}          & \underline{65.44}          & \multicolumn{1}{c|}{\underline{59.86}}          & 48.48          & \multicolumn{1}{c|}{\underline{46.96}}          & 62.55                        & \multicolumn{1}{c|}{38.48}       & 50.4                         & \multicolumn{1}{c|}{34.22}       \\
\multicolumn{1}{|c}{\textbf{def + ocr}} & \multicolumn{1}{c|}{\textbf{vn}}  & 50.21                                & \multicolumn{1}{c|}{34.87}          & \underline{51.49}          & \multicolumn{1}{c|}{\underline{38.19}}          & 64.13          & \multicolumn{1}{c|}{52.84}          & \colorbox[HTML]{DCDCDC}{51.49}          & \multicolumn{1}{c|}{\colorbox[HTML]{DCDCDC}{44.12}}          & \underline{62.77}                        & \multicolumn{1}{c|}{38.92}       & \underline{51.32}                        & \multicolumn{1}{c|}{\underline{37.32}}       \\
\multicolumn{1}{|c}{\textbf{vn}}        & \multicolumn{1}{c|}{\textbf{ex}}  & {\colorbox[HTML]{DCDCDC}{48.38}}         & \multicolumn{1}{c|}{\colorbox[HTML]{DCDCDC}{38.06}}          & 50.35          & \multicolumn{1}{c|}{35.2}           & 46.63          & \multicolumn{1}{c|}{41.27}          & 44.84          & \multicolumn{1}{c|}{44.21}          & {38.78} & \multicolumn{1}{c|}{32.18}       & {50.2}  & \multicolumn{1}{c|}{34.13}       \\
\multicolumn{1}{|c}{\textbf{def}}       & \multicolumn{1}{c|}{\textbf{ex}}  & {49.68}         & \multicolumn{1}{c|}{33.19}          & \colorbox[HTML]{DCDCDC}{51.43}          & \multicolumn{1}{c|}{\colorbox[HTML]{DCDCDC}{49.85}}          & \colorbox[HTML]{DCDCDC}{46.88}          & \multicolumn{1}{c|}{\colorbox[HTML]{DCDCDC}{46.77}}          & \colorbox[HTML]{DCDCDC}{50.35}          & \multicolumn{1}{c|}{\colorbox[HTML]{DCDCDC}{50.27}}          & {\colorbox[HTML]{DCDCDC}{62.46}} & \multicolumn{1}{c|}{\colorbox[HTML]{DCDCDC}{40.35}}       & {\colorbox[HTML]{DCDCDC}{49.71}} & \multicolumn{1}{c|}{\colorbox[HTML]{DCDCDC}{34.67}}       \\
\multicolumn{1}{|c}{\textbf{ocr}}       & \multicolumn{1}{c|}{\textbf{ex}}  & 49.12                                & \multicolumn{1}{c|}{34.41}          & \colorbox[HTML]{DCDCDC}{47.39}          & \multicolumn{1}{c|}{\colorbox[HTML]{DCDCDC}{47.34}}          & \colorbox[HTML]{DCDCDC}{65.42}          & \multicolumn{1}{c|}{\colorbox[HTML]{DCDCDC}{55.19}}          & \colorbox[HTML]{DCDCDC}{49.5}           & \multicolumn{1}{c|}{\colorbox[HTML]{DCDCDC}{45.17}}          & 61.66                        & \multicolumn{1}{c|}{39.6}        & 49.89                        & \multicolumn{1}{c|}{37.13}       \\
\multicolumn{1}{|c}{\textbf{def + ocr}} & \multicolumn{1}{c|}{\textbf{ex}}  & {\underline{53.06}}         & \multicolumn{1}{c|}{\underline{44.37}}          & \colorbox[HTML]{DCDCDC}{54.39}          & \multicolumn{1}{c|}{\colorbox[HTML]{DCDCDC}{52.52}}          & \colorbox[HTML]{DCDCDC}{65.6}           & \multicolumn{1}{c|}{\colorbox[HTML]{DCDCDC}{51.25}}          & \colorbox[HTML]{DCDCDC}{54.09}          & \multicolumn{1}{c|}{\colorbox[HTML]{DCDCDC}{49.82}}          & {61.57} & \multicolumn{1}{c|}{\underline{40.84}}       & {50.33} & \multicolumn{1}{c|}{35.37}       \\ \hline
                                        &                                   & \multicolumn{12}{c}{\scriptsize\textbf{\sysB{} \textsc{Flan-T5-xl}}}                                                                                                                                                                                                                                                                                                                                   \\ \hline
\multicolumn{1}{|c}{\textbf{vn}}        & \multicolumn{1}{c|}{\textbf{vn}}  & 50.2                                 & \multicolumn{1}{c|}{33.78}          & 56.67          & \multicolumn{1}{c|}{48.6}           & 64.12          & \multicolumn{1}{c|}{41.88}          & 51.27          & \multicolumn{1}{c|}{36.32}          & 62.68                        & \multicolumn{1}{c|}{38.53}       & 50.1                         & \multicolumn{1}{c|}{33.73}       \\
\multicolumn{1}{|c}{\textbf{def}}       & \multicolumn{1}{c|}{\textbf{vn}}  & 50.2                                 & \multicolumn{1}{c|}{35.13}          & \underline{59.9}           & \multicolumn{1}{c|}{\underline{56.67}}          & 64.97          & \multicolumn{1}{c|}{39.38}          & 51.27          & \multicolumn{1}{c|}{33.89}          & \underline{62.82}                        & \multicolumn{1}{c|}{38.94}       & 52.1                         & \multicolumn{1}{c|}{38}         \\
\multicolumn{1}{|c}{\textbf{ocr}}       & \multicolumn{1}{c|}{\textbf{vn}}  & 51.2                                 & \multicolumn{1}{c|}{39.22}          & 55.9           & \multicolumn{1}{c|}{47.75}          & 65.16          & \multicolumn{1}{c|}{40.21}          & 51.27          & \multicolumn{1}{c|}{33.89}          & 62.39                        & \multicolumn{1}{c|}{39.12}       & 51.9                         & \multicolumn{1}{c|}{37.89}       \\
\multicolumn{1}{|c}{\textbf{def + ocr}} & \multicolumn{1}{c|}{\textbf{vn}}  & \underline{52.6}                                 & \multicolumn{1}{c|}{\underline{42.33}}          & 52.1           & \multicolumn{1}{c|}{39.59}          & \underline{65.54}          & \multicolumn{1}{c|}{\underline{42.52}}          & \underline{51.98}          & \multicolumn{1}{c|}{\underline{36.68}}          & 62.62                        & \multicolumn{1}{c|}{39.21}       & \underline{52.2}                         & \multicolumn{1}{c|}{\underline{38.3}}        \\
\multicolumn{1}{|c}{\textbf{vn}}        & \multicolumn{1}{c|}{\textbf{ex}}  & {\colorbox[HTML]{DCDCDC}{60.78}}         & \multicolumn{1}{c|}{\colorbox[HTML]{DCDCDC}{60.76}}          & \colorbox[HTML]{DCDCDC}{55.34}          & \multicolumn{1}{c|}{\colorbox[HTML]{DCDCDC}{50.7}}           & \colorbox[HTML]{DCDCDC}{60.19}          & \multicolumn{1}{c|}{\colorbox[HTML]{DCDCDC}{41.53}}          & \colorbox[HTML]{DCDCDC}{49.12}          & \multicolumn{1}{c|}{\colorbox[HTML]{DCDCDC}{46.76}}          & {\colorbox[HTML]{DCDCDC}{41.39}} & \multicolumn{1}{c|}{\colorbox[HTML]{DCDCDC}{37.21}}       & {\colorbox[HTML]{DCDCDC}{48.33}} & \multicolumn{1}{c|}{\colorbox[HTML]{DCDCDC}{33.14}}       \\
\multicolumn{1}{|c}{\textbf{def}}       & \multicolumn{1}{c|}{\textbf{ex}}  & {51}            & \multicolumn{1}{c|}{40.86}          & \colorbox[HTML]{DCDCDC}{53.14}          & \multicolumn{1}{c|}{\colorbox[HTML]{DCDCDC}{34.92}}          & 64.2           & \multicolumn{1}{c|}{39.1}           & 51.46          & \multicolumn{1}{c|}{33.98}          & {\colorbox[HTML]{DCDCDC}{36.87}} & \multicolumn{1}{c|}{\colorbox[HTML]{DCDCDC}{29.08}}       & {50.1}  & \multicolumn{1}{c|}{33.74}       \\
\multicolumn{1}{|c}{\textbf{ocr}}       & \multicolumn{1}{c|}{\textbf{ex}}  & \colorbox[HTML]{DCDCDC}{60}                                   & \multicolumn{1}{c|}{\colorbox[HTML]{DCDCDC}{58.17}}          & \colorbox[HTML]{DCDCDC}{61.13}          & \multicolumn{1}{c|}{\colorbox[HTML]{DCDCDC}{59.39}}          & \colorbox[HTML]{DCDCDC}{61.7}           & \multicolumn{1}{c|}{\colorbox[HTML]{DCDCDC}{38.16}}          & \colorbox[HTML]{DCDCDC}{49.33}          & \multicolumn{1}{c|}{\colorbox[HTML]{DCDCDC}{33.03}}          & \colorbox[HTML]{DCDCDC}{41.47}                        & \multicolumn{1}{c|}{\colorbox[HTML]{DCDCDC}{36.08}}       & {50}    & \multicolumn{1}{c|}{33.33}       \\
\multicolumn{1}{|c}{\textbf{def + ocr}} & \multicolumn{1}{c|}{\textbf{ex}}  & {\colorbox[HTML]{DCDCDC}{57.94}}         & \multicolumn{1}{c|}{\colorbox[HTML]{DCDCDC}{55.58}}          & 55.01          & \multicolumn{1}{c|}{45.67}          & 65.04          & \multicolumn{1}{c|}{39.41}          & 51.56          & \multicolumn{1}{c|}{34.02}          & {62.61} & \multicolumn{1}{c|}{\underline{47.97}}       & {49.9}  & \multicolumn{1}{c|}{33.29}       \\ \hline
                                        &                                   & \multicolumn{12}{c}{\scriptsize\textbf{\sysGPT{}}}                                                                                                                                                                                                                                                                                                                                           \\ \hline
\multicolumn{1}{|c}{\textbf{vn}}        & \multicolumn{1}{c|}{\textbf{vn}}  & 69.2                                 & \multicolumn{1}{c|}{68.02}          & 79.3           & \multicolumn{1}{c|}{78.84}          & 71.47          & \multicolumn{1}{c|}{69.54}          & 67.04          & \multicolumn{1}{c|}{66.99}          & 64.84                        & \multicolumn{1}{c|}{\underline{\textbf{59.07}}}       & \underline{\textbf{65}}                          & \multicolumn{1}{c|}{\underline{\textbf{64.24}}}       \\
\multicolumn{1}{|c}{\textbf{def}}       & \multicolumn{1}{c|}{\textbf{vn}}  & 70.4                                 & \multicolumn{1}{c|}{69.32}          & 79.5           & \multicolumn{1}{c|}{78.95}          & 72.32          & \multicolumn{1}{c|}{71.71}          & 63.66          & \multicolumn{1}{c|}{62.52}          & 62.31                        & \multicolumn{1}{c|}{55.56}       & 62.6                         & \multicolumn{1}{c|}{61.64}       \\
\multicolumn{1}{|c}{\textbf{ocr}}       & \multicolumn{1}{c|}{\textbf{vn}}  & 70                                   & \multicolumn{1}{c|}{69.5}           & \underline{\textbf{83.6}}  & \multicolumn{1}{c|}{\underline{\textbf{83.54}}} & 67.8           & \multicolumn{1}{c|}{63.46}          & 64.23          & \multicolumn{1}{c|}{63.44}          & \underline{\textbf{66.95} }                       & \multicolumn{1}{c|}{57.87}       & 61                          & \multicolumn{1}{c|}{58.74}       \\
\multicolumn{1}{|c}{\textbf{def + ocr}} & \multicolumn{1}{c|}{\textbf{vn}}  & 72.2                                 & \multicolumn{1}{c|}{71.16}          & 80.3           & \multicolumn{1}{c|}{79.96}          & 71.47          & \multicolumn{1}{c|}{70.86}          & 62.54          & \multicolumn{1}{c|}{62.15}          & 66.1                         & \multicolumn{1}{c|}{56.94}       & 61.6                         & \multicolumn{1}{c|}{59.57}       \\
\multicolumn{1}{|c}{\textbf{vn}}        & \multicolumn{1}{c|}{\textbf{ex}}  & {70.4}          & \multicolumn{1}{c|}{69.74}          & 82.9           & \multicolumn{1}{c|}{82.78}          & 69.77          & \multicolumn{1}{c|}{66.21}          & \underline{\textbf{69.3}}  & \multicolumn{1}{c|}{\underline{\textbf{69.29}}} & {65.26} & \multicolumn{1}{c|}{56.01}       & {60.8}  & \multicolumn{1}{c|}{59.17}       \\
\multicolumn{1}{|c}{\textbf{def}}       & \multicolumn{1}{c|}{\textbf{ex}}  & {72.8}          & \multicolumn{1}{c|}{72.35}          & 82.03          & \multicolumn{1}{c|}{81.86}          & 72.03          & \multicolumn{1}{c|}{71.06}          & 65.35          & \multicolumn{1}{c|}{64.59}          & {60.76} & \multicolumn{1}{c|}{55.22}       & {64.2}  & \multicolumn{1}{c|}{63.5}        \\
\multicolumn{1}{|c}{\textbf{ocr}}       & \multicolumn{1}{c|}{\textbf{ex}}  & 69.8                                 & \multicolumn{1}{c|}{69.68}          & 83.03          & \multicolumn{1}{c|}{83.03}          & 69.49          & \multicolumn{1}{c|}{62.82}          & 64.51          & \multicolumn{1}{c|}{62.66}          & 65.54                        & \multicolumn{1}{c|}{57.83}       & {61.4}  & \multicolumn{1}{c|}{59.47}       \\
\multicolumn{1}{|c}{\textbf{def + ocr}} & \multicolumn{1}{c|}{\textbf{ex}}  & {\underline{\textbf{73.4}}} & \multicolumn{1}{c|}{\underline{\textbf{73.1}}}  & 82.54          & \multicolumn{1}{c|}{82.52}          & \underline{\textbf{74.58}} & \multicolumn{1}{c|}{\underline{\textbf{72.82}}} & 63.94          & \multicolumn{1}{c|}{63.91}          & {65.96} & \multicolumn{1}{c|}{59}         & {61.4}  & \multicolumn{1}{c|}{59.47}       \\ \hline
\end{tabular}
\caption{\footnotesize\textsc{\textbf{Overall Results:}} \colorbox[HTML]{DCDCDC}{Greyed} out cells signify ambiguity above assigned threshold. Best \textit{(model, prompt)} for each \textit{(model, dataset)} is \underline{underlined}. Best \textit{(model, prompt)} for each dataset is marked \underline{\textbf{bold \& underlined}}; across open source models it is marked \textbf{bold}.}
\label{tab:all_prompt_model_results}
\end{table*}

\begin{table}[!th]
\centering
\scriptsize
\setlength{\tabcolsep}{1mm}
\begin{tabular}{|cc|cccccc|}
\hline
\multicolumn{2}{|c|}{\textbf{Strategy}} & \multicolumn{6}{c|}{\textbf{Models}}                                                                                                                                                       \\ \hline
\textbf{in}           & \textbf{out}    & \multicolumn{1}{c|}{\textbf{ID(9B)}} & \multicolumn{1}{c|}{\textbf{LV(13B)}}            & \multicolumn{1}{c|}{\textbf{LV(7B)}}    & \multicolumn{1}{c|}{\textbf{IB(V)}} & \multicolumn{1}{c|}{\textbf{IV(F)}} & \multicolumn{1}{c|}{\textbf{\sysGPT{}}}\\ \hline
\textbf{vn}           & \textbf{vn}     & \multicolumn{1}{c|}{41.11}      & \multicolumn{1}{c|}{\colorbox[HTML]{DCDCDC}{\textit{52.1}}}          & \multicolumn{1}{c|}{\colorbox[HTML]{DCDCDC}{38.71}}       & \multicolumn{1}{c|}{\colorbox[HTML]{DCDCDC}{38.07}}            & \multicolumn{1}{c|}{40.2}             & \multicolumn{1}{c|}{\underline{68.82}}          \\
\textbf{def}          & \textbf{vn}     & \multicolumn{1}{c|}{\colorbox[HTML]{DCDCDC}{36.03}}         & \multicolumn{1}{c|}{\textit{{57.12}}}          & \multicolumn{1}{c|}{\colorbox[HTML]{DCDCDC}{45.85}}       & \multicolumn{1}{c|}{\colorbox[HTML]{DCDCDC}{39.24}}            & \multicolumn{1}{c|}{42.95}            & \multicolumn{1}{c|}{\underline{67.69}}          \\
\textbf{ocr}          & \textbf{vn}     & \multicolumn{1}{c|}{51.18}      & \multicolumn{1}{c|}{\colorbox[HTML]{DCDCDC}{\textit{53.74}}}           & \multicolumn{1}{c|}{\colorbox[HTML]{DCDCDC}{44.16}}       & \multicolumn{1}{c|}{38.93}         & \multicolumn{1}{c|}{41.05}            & \multicolumn{1}{c|}{\underline{68.36}}          \\
\textbf{def + ocr}    & \textbf{vn}     & \multicolumn{1}{c|}{38.78}       & \multicolumn{1}{c|}{\textit{{\underline{58.45}}}} & \multicolumn{1}{c|}{\colorbox[HTML]{DCDCDC}{52.39}}       & \multicolumn{1}{c|}{\colorbox[HTML]{DCDCDC}{39.86}}            & \multicolumn{1}{c|}{39.72}            & \multicolumn{1}{c|}{\underline{68.12}}          \\
\textbf{vn}           & \textbf{ex}     & \multicolumn{1}{c|}{38.77}      & \multicolumn{1}{c|}{\textit{\colorbox[HTML]{DCDCDC}{53.35}}}          & \multicolumn{1}{c|}{\colorbox[HTML]{DCDCDC}{43.43}}       & \multicolumn{1}{c|}{\colorbox[HTML]{DCDCDC}{36.4}}            & \multicolumn{1}{c|}{\colorbox[HTML]{DCDCDC}{45.44}}            & \multicolumn{1}{c|}{\underline{68.74}}             \\
\textbf{def}          & \textbf{ex}     & \multicolumn{1}{c|}{35.26}      & \multicolumn{1}{c|}{{\textit{44.38}}}          & \multicolumn{1}{c|}{\colorbox[HTML]{DCDCDC}{42.14}}       & \multicolumn{1}{c|}{\colorbox[HTML]{DCDCDC}{42.94}}            & \multicolumn{1}{c|}{\colorbox[HTML]{DCDCDC}{34.74}}            & \multicolumn{1}{c|}{\underline{69.34}}             \\
\textbf{ocr}          & \textbf{ex}     & \multicolumn{1}{c|}{48.81}      & \multicolumn{1}{c|}{\textit{{54.36}}}          & \multicolumn{1}{c|}{41.59}       & \multicolumn{1}{c|}{\colorbox[HTML]{DCDCDC}{42.94}}            & \multicolumn{1}{c|}{\colorbox[HTML]{DCDCDC}{45.62}}            & \multicolumn{1}{c|}{\underline{68.19}}             \\
\textbf{def + ocr}    & \textbf{ex}     & \multicolumn{1}{c|}{37.34}      & \multicolumn{1}{c|}{45.84}                & \multicolumn{1}{c|}{\textit{{53.39}}} & \multicolumn{1}{c|}{\colorbox[HTML]{DCDCDC}{45.98}}            & \multicolumn{1}{c|}{\colorbox[HTML]{DCDCDC}{43.93}}            & \multicolumn{1}{c|}{\textbf{\underline{69.95}}}             \\ \hline
\multicolumn{2}{|c|}{\textbf{Mean}}     & \multicolumn{1}{c|}{\colorbox[HTML]{DCDCDC}{40.91}}      & \multicolumn{1}{c|}{\colorbox[HTML]{DCDCDC}{\textit{\underline{52.42}}}}                & \multicolumn{1}{c|}{\colorbox[HTML]{DCDCDC}{45.21}} & \multicolumn{1}{c|}{\colorbox[HTML]{DCDCDC}{40.54}}            & \multicolumn{1}{c|}{\colorbox[HTML]{DCDCDC}{41.71}}            & \multicolumn{1}{c|}{\textbf{\underline{68.65}}}             \\
\multicolumn{2}{|c|}{\textbf{(Std. Dev.)}}     & \multicolumn{1}{c|}{\colorbox[HTML]{DCDCDC}{(5.54)}}      & \multicolumn{1}{c|}{\colorbox[HTML]{DCDCDC}{(4.65)}}                & \multicolumn{1}{c|}{\colorbox[HTML]{DCDCDC}{(4.85)}} & \multicolumn{1}{c|}{\colorbox[HTML]{DCDCDC}{\underline{\textit{(2.94)}}}}            & \multicolumn{1}{c|}{\colorbox[HTML]{DCDCDC}{(3.37)}}            & \multicolumn{1}{c|}{\textbf{\underline{(0.68)}}}             \\ \hline
\end{tabular}
\caption{\footnotesize\textsc{\textbf{Leaderboard and Stability:}} Weighted macro-F1 score across 6 datasets with mean + (standard deviation) along promt variants. Overall best macro-F1 score and best mean + (standard deviation) are marked \underline{\textbf{bold and underlined}}; across open source models are \underline{\textit{italicized and underlined}}. Best macro-F1 scores across each prompt variant are \underline{underlined}; for open source models are \textit{italicized}. \colorbox[HTML]{DCDCDC}{Greyed} out cells signify ambiguity above assigned threshold; similar to Table~\ref{tab:all_prompt_model_results}. \textbf{ID(9B)}: \sysI{} 9B, \textbf{LV(13B), LV(7B)}: \sysL{} 13B, 7B, \textbf{I-BLIP V}: \sysB{} \textsc{Vicuna 7B}, \textbf{I-BLIP F}: \sysB{} \textsc{Flan-T5-xl}.}
\label{tab:weighted_prompt_model_output}
\end{table}

In this section, we present the results of our experiments. In Table~\ref{tab:all_prompt_model_results} we show the results for the six datasets across the six models. Each block in the table corresponds to a particular (model, dataset) combination and covers the results for eight prompt pattern combinations. Since we use the generation capability of VLMs for prediction, we observe that in some prompt variants, certain (model, prompt) combinations did not classify the input meme amongst the \texttt{list\_of\_labels} and diplomatically bypassed the query with an irrelevant answer. This led to a decrease in support to infer the results accurately grounded on correct labels in the dataset. In the table, we have greyed out the cases which did not generate a correct label for at least 90\% of the data points. Examples of ambiguous outputs are provided in Appendix\ref{sec:outputs_extended}.\\

\noindent\textbf{Overall results\label{subsection:overall_results}}: From Table~\ref{tab:all_prompt_model_results}, we observe that \sysB{} models are not able to correctly predict the labels out of \texttt{list\_of\_labels} even for English datasets and generate ambiguous answers for quite a large number of prompt variants. Their generated output did not conform with the expected output format specified in input prompt. We also observe that \sysI{} performs best with only OCR as input. \sysL{} 7B, works best with explanation as output, when the input prompt was definition \& OCR text. Overall, \sysGPT{}  emerged to be the best model and  \sysL{} 13B the best open-source model. While \sysL{} 13B works best with OCR \& definition as input and vanilla output, for \sysGPT{} the best prompt variant is not conclusive as per Table~\ref{tab:all_prompt_model_results}. However, we observe that the variation of scores is quite less in \sysGPT{} compared to open source models and therefore, in the upcoming subsection we formulate a \textit{Leaderboard} to further understand, quantify and provide in-depth conclusions of our obtained results.\\ \textbf{Multilingual capability}: We observe that even \sysL{} 7B \& 13B models generate ambiguous outputs for BHM and \textsc{HinGlish} datasets; indicating their monolingual limitation. For all models, including \sysGPT{}, the metrics degrade considerably compared to the results on English datasets. This demonstrates the limitations of VLMs in handling multilingual memes. As expected, \sysGPT{} outperforms all models in terms of metrics and does not generate any ambiguous output.

\noindent\textbf{Leaderboard: \label{subsection:leaderboard}}Since engineering solutions are always in the \textit{`quest for the best'}, we propose a quantitative metric to organize the (model, prompt) combinations into a leaderboard. The idea is that the top combinations on this leaderboard should generalize well across the six datasets combined (and eventually  across 4 different dimensions of toxicity, i.e., hate, misogyny, harm and offense and 3 different languages, i.e., English, Bangla and \textsc{HinGlish}). For each prompt variant considered over all models, we calculate a weighted average macro-F1 score depending on the number of samples in each of the datasets by the formulation:
$\frac{\sum\limits_{\mathcal{D}} (f_\mathcal{D})*|\mathcal{D}|} {\sum\limits_{\mathcal{D}}|\mathcal{D}|}$. Here $f_\mathcal{D}$ is the macro-F1 for the dataset $\mathcal{D}$. After calculating the weighted average macro-F1 score, we also calculate mean and standard deviation of each model across all prompt variants to obtain the overall performance and stability of the models. The results are shown Table~\ref{tab:weighted_prompt_model_output}.

Based on the above results, we conclude \sysGPT{} to be the best model with definition \& OCR as input and explanation as output. \sysL{} 13B is the best open-source model with definition \& OCR text as input and vanilla as output. Moreover, \sysGPT{} is the best model for all prompt variants taken individually and \sysL{} 13B outperforms other open-source models in 7 of the 8 prompt variants and lags behind \sysL{} 7B variant for the setting with definition \& OCR as input and explanation as output.\\\textbf{Stability}: Further, we also observe that \sysGPT{} is not only the best model across all prompt variants, but is also the most stable model with least standard deviation across different prompt variants. Amongst open-source models, \sysL{} 13B model has the best mean.

\noindent\textbf{Baselines:} As per the discussion in `Models' section, we present the results in Table~\ref{tab:baseline_results} on zero-shot and FHM fine-tuned versions of the baseline. Note that it is done to be very near to zero-shot evaluation. \sysGPT{} outperforms these baselines even without any fine-tuning. Notably, \sysL{} 13B version also outperforms these baselines across Macro-F1 score. This raises a deep concern about the generalization capability of previous works.

\begin{table}[!t]
\centering
\scriptsize
\begin{tabular}{c|c|cc}
\textbf{Datasets}                & \textbf{Metrics} & \textbf{zero shot} & \textbf{fine-tuned} \\ \hline
\multirow{2}{*}{\textbf{FHM}}    & \textbf{acc}     & 52.13 {\scriptsize\textit{(0.95)}}              & 60.67 {\scriptsize\textit{(1.33)}}              \\
                                 & \textbf{mf1}     & 47.11 {\scriptsize\textit{(0.97)}}              & 57.27 {\scriptsize\textit{(1.53)}}              \\ \cline{1-1}
\multirow{2}{*}{\textbf{MAMI}}   & \textbf{acc}     & 50.43 {\scriptsize\textit{(0.51)}}             & 59.57 {\scriptsize\textit{(0.32)}}              \\
                                 & \textbf{mf1}     & 43.88 {\scriptsize\textit{(0.47)}}             & 56.58 {\scriptsize\textit{(0.43)}}              \\ \cline{1-1}
\multirow{2}{*}{\textbf{HARM-C}} & \textbf{acc}     & 57.72 {\scriptsize\textit{(0.99)}}             & 61.21 {\scriptsize\textit{(1.27)}}              \\
                                 & \textbf{mf1}     & 46.57 {\scriptsize\textit{(1.52)}}             & 44.72 {\scriptsize\textit{(1.34)}}              \\ \cline{1-1}
\multirow{2}{*}{\textbf{HARM-P}}          & \textbf{acc}     & 52.14 {\scriptsize\textit{(1.73)}}             & 54.76 {\scriptsize\textit{(0.65)}}              \\
                                 & \textbf{mf1}     & 44.67 {\scriptsize\textit{(2.65)}}             & 44.81 {\scriptsize\textit{(1.51)}}              \\ \cline{1-1}
\multirow{2}{*}{\textbf{BHM}}             & \textbf{acc}     & 57.48 {\scriptsize\textit{(1.38)}}             & 60.62 {\scriptsize\textit{(1.02)}}              \\
                                 & \textbf{mf1}     & 44.92 {\scriptsize\textit{(2.1)}}             & 50.28 {\scriptsize\textit{(1.73)}}              \\ \cline{1-1}
\multirow{2}{*}{\textbf{\textsc{HinGlish}}}        & \textbf{acc}     & 50.2 {\scriptsize\textit{(0.6)}}              & 55.13 {\scriptsize\textit{(0.23)}}              \\
                                 & \textbf{mf1}     & 38.55 {\scriptsize\textit{(0.68)}}             & 49.03 {\scriptsize\textit{(0.41)}}             
\end{tabular}
\caption{\footnotesize\textbf{\textsc{Baselines:}} Mean with standard deviation in \textit{(parenthesis)}. We have discussed details in Section \textbf{`Models'}.}
\label{tab:baseline_results}
\end{table}
%%%%%%%%%%%%%%%%%%%%%%%%%%%%%%%%%%%%%%%%%%%%%%%%%%%%%%%%%%%%%%%%%%%%%%%%
\section{Error analysis}
In the previous section we concluded that \sysGPT{} (with definition + OCR text as input and explanation as output) is the best model and \sysL{} 13B (with definition + OCR text as input and vanilla as output) is the best model amongst chosen open source models. Since it is practically infeasible to study error analysis over all models and prompt variants, we therefore investigate the cases of misclassification for these two best models (one proprietary and one open source based on the \textit{Leaderboard}) with their best prompt variants. We comprehensively evaluate \sysGPT{} for a total of 828 misclassified memes: 124- FHM, 149- MAMI, 195- HARM P+C, 205- BHM and 155- \textsc{HinGlish} datasets  and \sysL{} 13B for a total of 1184 misclassified memes; 202- FHM, 321- MAMI, 276- HARM P+C, 216- BHM and 169- \textsc{HinGlish} datasets. In particular we attempt to obtain an explanation of \textit{\textbf{parts in the meme}} that confuses the model and leads to mispredictions (first sub-section). Further, we induce a \textbf{\textit{typology of the error cases}} to systematically organise the vulnerable points of the model (second sub-section).

\subsection{Occlusion based result interpretation}

\label{subsection:8.1}
\begin{table}[!ht]
\centering
\scriptsize
\begin{tabular}{c|c|c|c}
\textbf{Dataset}                                                                      & \textbf{Model}     & \textbf{CASES: 1, 2}  & \textbf{CASES: 3, 4}  \\ \hline
\multirow{2}{*}{\textbf{FHM}}                                                & \textbf{\sysGPT{}}    & \textbf{15.32, 55.65} & \textbf{8.06, 20.97}  \\
                                                                             & \textbf{LV-13B} & \textbf{14.36, 26.73} & \textbf{20.3, 38.61}  \\ \hline
\multirow{2}{*}{\textbf{MAMI}}                                               & \textbf{\sysGPT{}}    & \textbf{17.45, 42.95} & \textbf{14.77, 24.83} \\
                                                                             & \textbf{LV-13B} & 20.87, 15.89 & 38.32, 24.92 \\ \hline
\multirow{2}{*}{\textbf{\begin{tabular}[c]{@{}c@{}}HARM\\ P+C\end{tabular}}} & \textbf{\sysGPT{}}    & \textbf{18.97, 33.85} & \textbf{12.82, 34.36} \\
                                                                             & \textbf{LV-13B} & 34.78, 30.07 & 20.29, 14.86 \\ \hline
\multirow{2}{*}{\textbf{\begin{tabular}[c]{@{}c@{}}BHM\end{tabular}}} & \textbf{\sysGPT{}}    & 12.68, 9.76 & \textbf{26.83, 50.73} \\
                                                                             & \textbf{LV-13B} & 7.87, 2.31 & \textbf{35.19, 54.63} \\ \hline
\multirow{2}{*}{\textbf{\begin{tabular}[c]{@{}c@{}}\textsc{HinGlish}\end{tabular}}} & \textbf{\sysGPT{}}    & 13.55, 7.1 & \textbf{34.19, 45.16} \\
                                                                             & \textbf{LV-13B} & 18.93, 5.33 & 42.60, 33.14
\end{tabular}
\caption{\footnotesize\textbf{\textsc{Case Distribution and Rigidness:}} Percentage distribution of misclassified samples from each dataset for all 4 CASES studied under occlusion based interpretability. All numbers are in percentage (\%). Wherever CASE 2 or 4 are higher than CASE 1 or 3 respectively, the entries are marked in \textbf{bold} as a measure of \textbf{rigidness} of models towards occlusion.}
\label{tab:percentage_distribution_slic_superpixels}
\end{table}

\begin{table}[!ht]
\centering
\scriptsize
\label{tab:slic_quantized_result}
\renewcommand{\arraystretch}{1}
\setlength{\tabcolsep}{0.5mm}
\begin{tabular}{c|c|c|c|c|c}
\textbf{Dataset} & \textbf{Model} & \textbf{CASE 1} & \textbf{CASE 2} & \textbf{CASE 3} & \textbf{CASE 4} \\ \hline
\multirow{2}{*}{\textbf{FHM}} & \sysGPT{} & \begin{tabular}[c]{@{}c@{}}NHM: \textbf{47.37}\\WA: \textbf{36.84}\\SI: \textbf{26.32} \end{tabular} & \begin{tabular}[c]{@{}c@{}}WA:\textbf{72.46}\\NHM: \textbf{18.84} \end{tabular} & \begin{tabular}[c]{@{}c@{}}MCVG: \textbf{50}\\URET: \textbf{30} \end{tabular} & IM: \textbf{73.08} \\ \cline{2-6} 
 & LV-13B & \begin{tabular}[c]{@{}c@{}}WA: \textbf{41.38}\\SI: \textbf{24.14}\\NHM: \textbf{24.14}  \end{tabular} & \begin{tabular}[c]{@{}c@{}}WA: \textbf{62.96}\\ NHM: \textbf{24.07} \end{tabular} & \begin{tabular}[c]{@{}c@{}}MCVG: \textbf{41.46}\\URET: \textbf{19.51}\\SI: \textbf{12.2}\end{tabular} & IM: \textbf{74.36} \\ \hline
\multirow{2}{*}{\textbf{MAMI}} & \sysGPT{} & \begin{tabular}[c]{@{}c@{}}WA: \textbf{69.23}\\PFW: \textbf{34.62} \end{tabular} & WA: \textbf{64.06} & NV: \textbf{54.54} & IM: \textbf{70.27}\\ \cline{2-6} 
 & LV-13B & \begin{tabular}[c]{@{}c@{}}WA: \textbf{44.78}\\ SI: \textbf{19.4}\\PFW: \textbf{17.91} \end{tabular} & WA: \textbf{64.71} & NV: \textbf{47.15} & IM: \textbf{56.25}  \\ \hline
\multirow{2}{*}{\textbf{HARM CP}} & \sysGPT{} & \textbf{\textit{No conclusion}} & WA: \textbf{68.18} & URET: \textbf{24} & FC: \textbf{26.67} \\ \cline{2-6} 
 & LV-13B & \textbf{\textit{No conclusion}} & WA: \textbf{40.34} & URET: \textbf{30.36} & FC: \textbf{29.27} \\ \hline
\multirow{2}{*}{\textbf{BHM}} & \sysGPT{} & \begin{tabular}[c]{@{}c@{}}SI: \textbf{42.56}\end{tabular} & SI: \textbf{55} & SI: \textbf{60} & SI: \textbf{50.77}\\ \cline{2-6} 
 & LV-13B & \begin{tabular}[c]{@{}c@{}}SI: \textbf{44.89}\end{tabular} & SI: \textbf{69.65} & SI: \textbf{65.89} & SI: \textbf{59.7}  \\ \hline
\multirow{2}{*}{\textsc{\textbf{HinGlish}}} & \sysGPT{} & \begin{tabular}[c]{@{}c@{}}SI: \textbf{57.14} \end{tabular} & SI: \textbf{63.64} & \textbf{\textit{No conclusion}} & WA: \textbf{55.5} \\ \cline{2-6} 
 & LV-13B & \begin{tabular}[c]{@{}c@{}}SI: \textbf{54.54} \end{tabular} & SI: \textbf{62.51} & \textbf{\textit{No conclusion}} & WA: \textbf{56.25}
\end{tabular}
\caption{\footnotesize\textsc{\textbf{Class Based Error Typology:}} Distribution of various classes across misclassifications for \sysGPT{} \& \sysL{} 13B as per manual evaluation done on occlusion. All the numbers are in percentage (\%). Following are the 9 classes covered: \textbf{NHM}: Not hateful memes containing (i) common target words (i.e., Islam, white, migrants, etc.) (ii) image signifying these common targets or having some politician in it (iii) profane words, \textbf{WA}: Wrong annotation, \textbf{SI}: Stacked images, \textbf{MCVG}: Multiple color variations, grayscale image or multiple objects, \textbf{URET}: Lengthy, unreadable embedded text or having small font size, \textbf{IM}: Implicitly hate meme, \textbf{PFW}: Perturbed / animated faces of women, \textbf{NV}: Nudity or vulgarity in image or embedded text, \textbf{FC}: Fake conversation. \textbf{Note:} \textbf{HARM CP}: Harm C+P, \textbf{LV-13B:} \sysL{} 13B.}
\label{tab:slic_quantized_result}
\end{table}

Using the SLIC algorithm~\cite{slic2012}\footnote{Useful tutorial on SLIC algorithm -\\ 
 {\url{https://darshita1405.medium.com/superpixels-and-slic-6b2d8a6e4f08}}} we first segment the misclassfied memes into superpixels. The algorithm automatically segments the images into \textbf{\textit{5 - 12 superpixels}} depending on the size of the image. We control the size of each superpixel so that it is neither too small nor too big. Next the region circumscribing each of these superpixels are occluded one at a time by white patches and the models (i.e., \sysGPT{} and \sysL{} 13B) are queried again for its predictions with their corresponding best prompt variant (as per \textit{Leaderboard}).\\
\textbf{CASES:} Following are the cases in which we have broadly divided our occlusion study:
\textit{CASE 1}- Original meme misclassified as positive (i.e., hateful, misogynistic, harmful, or offensive corresponding to the dataset) and at least one occluded version resulted in the correct prediction (i.e., negative).
\textit{CASE 2}- Original meme misclassified as positive and none of the occluded versions resulted in the correct prediction .
\textit{CASE 3}- Original meme misclassified as negative and at least one occluded version resulted in the correct prediction (i.e., positive).
\textit{CASE 4}- Original meme misclassified as negative and none of the occluded versions resulted in the correct prediction.\\
\noindent\textbf{Rigidness}: In Table~\ref{tab:percentage_distribution_slic_superpixels}, we present the distribution for each of the cases. For \sysGPT{}, we observe that CASE 2 \& CASE 4 have higher values compared to CASE 1 and CASE 3 respectively. This points to a very interesting observation regarding the rigidness of \sysGPT{} in its classification decision. \sysL{} 13B on the other hand has similar behaviour for the FHM and BHM datasets. Overall these results show that the models which perform better are also more rigid in their generation and are comparatively less prone to a decision change based on the perturbations in the image. Decrease in rigidness on non-English datasets relates with the weaker performance of models on multilingual ability.\\
\noindent\textbf{Case by case study:} We perform further analysis and divide each case across 9 classes (refer Table~\ref{tab:slic_quantized_result}) and provide  representative examples in Table~\ref{tab:super_pixel_patch_results}. Based on these tables we present a {\textit{case-by-case manual analysis}} of the results. We employ three annotators to perform manual analysis of these occluded images. Out of them, two are NLP researchers having professional working experience on understanding hateful memes and the other one is an undergraduate student in his fourth year with major in Computer Science and has working experience in NLP. All the three annotators are from the authors' institute and are well aware of the topicality of discussion being conveyed in the memes. As we employ three annotators, we calculate Krippendorff's $\alpha$~\cite{castro-2017-fast-krippendorff} as a metric for inter-annotator agreement. We obtain high agreement values of 0.873 \& 0.875 for \sysGPT{} \& \sysL{} 13B, respectively. Note that we have added annotation guidlines in Appendix.
\begin{table*}[!t]
\centering
\scriptsize
\setlength{\tabcolsep}{0.05mm}
\begin{tabular}{c|c|cc|c}
\textbf{Dataset}                      & \textbf{Misclassified to}                      & \multicolumn{2}{c|}{\textbf{Change in prediction}} & \textbf{No change in prediction} \\ \hline
\multirow{4}{*}{\textbf{FHM}}         & \multirow{2}{*}{\textbf{Hateful}}      & \multicolumn{1}{c}{\includegraphics[width=0.22\columnwidth]
    {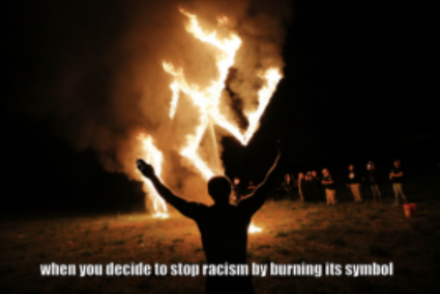}
    \includegraphics[width=0.2\columnwidth]
    {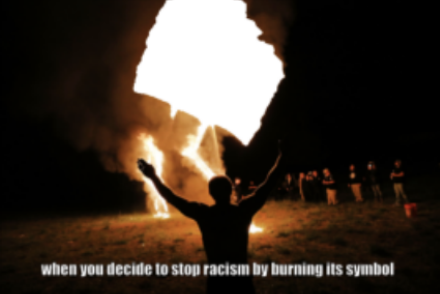}}                          & {\includegraphics[width=0.18\columnwidth]
    {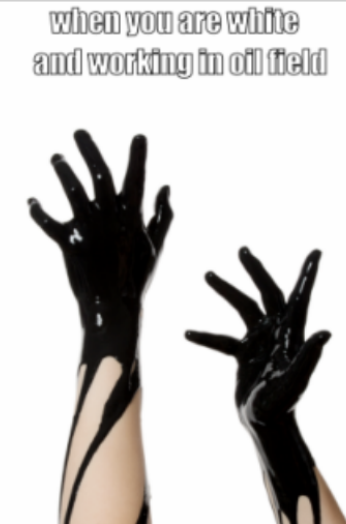}
    \includegraphics[width=0.18\columnwidth]
    {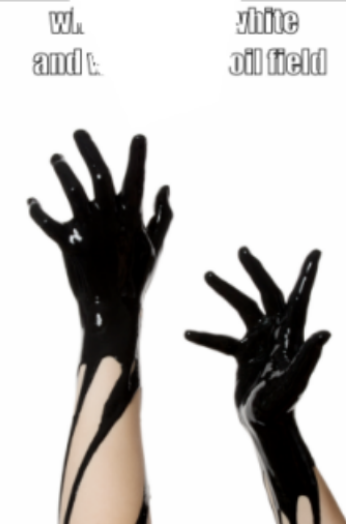}}                          &                                             {\includegraphics[width=0.16\columnwidth]
    {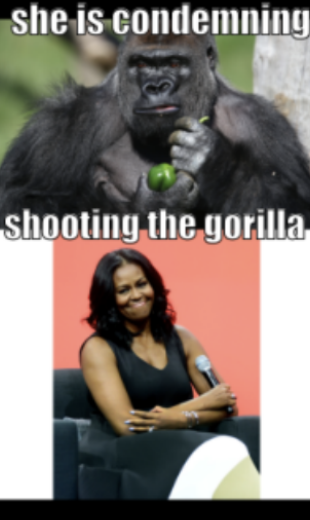}
    \includegraphics[width=0.2\columnwidth]
    {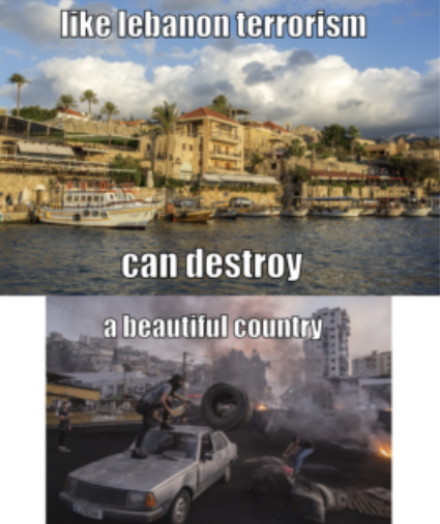}}                 \\ 
                                      &                                            & \multicolumn{2}{c|}{\textbf{CASE 1}}                                                   & \textbf{CASE 2}                                                      \\ \cline{2-5} 
                                      & \multirow{2}{*}{\textbf{Not-Hateful}}          & \multicolumn{1}{c}{\includegraphics[width=0.23\columnwidth]
    {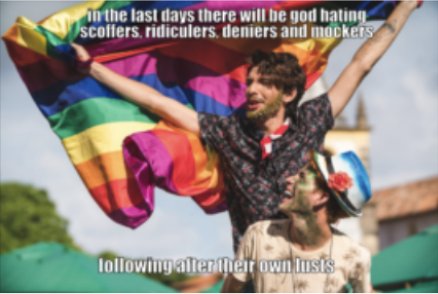}
    \includegraphics[width=0.2\columnwidth]
    {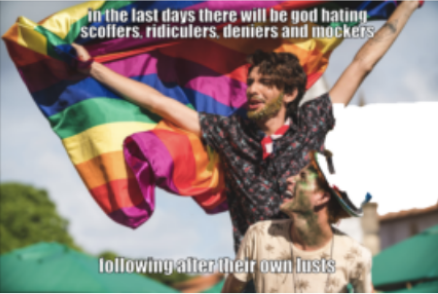}}                        & {\includegraphics[width=0.18\columnwidth]
    {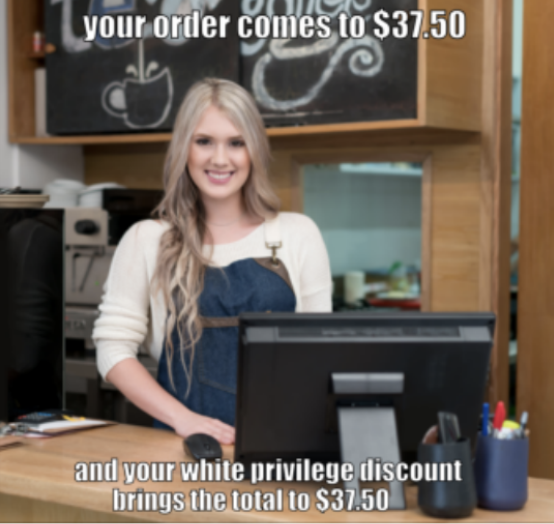}
    \includegraphics[width=0.18\columnwidth]
    {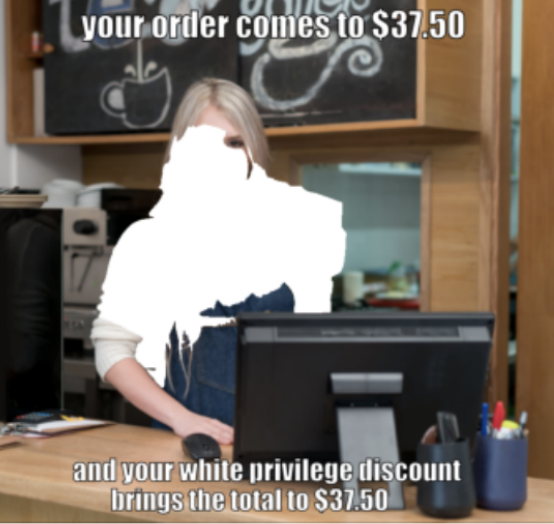}}                           &                                              {\includegraphics[width=0.2\columnwidth]
    {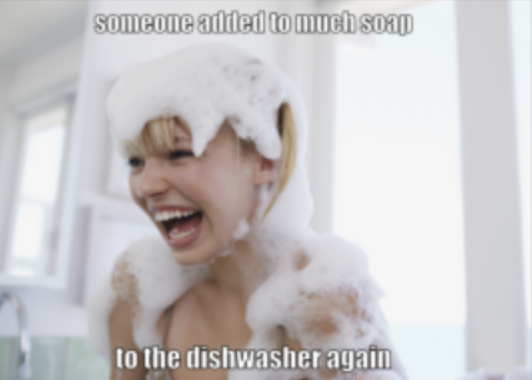}
    \includegraphics[width=0.2\columnwidth]
    {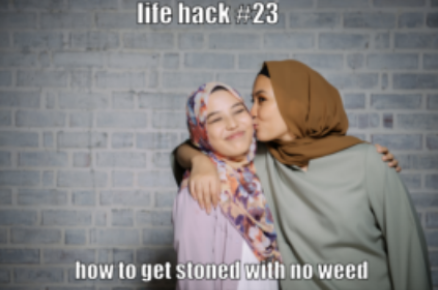}}                \\ 
                                      &                                            & \multicolumn{2}{c|}{\textbf{CASE 3}}                                                   & \textbf{CASE 4}                                                      \\ \hline
\multirow{4}{*}{\textbf{MAMI}}        & \multirow{2}{*}{\textbf{Misogynistic}} & \multicolumn{1}{c}{\includegraphics[width=0.22\columnwidth]
    {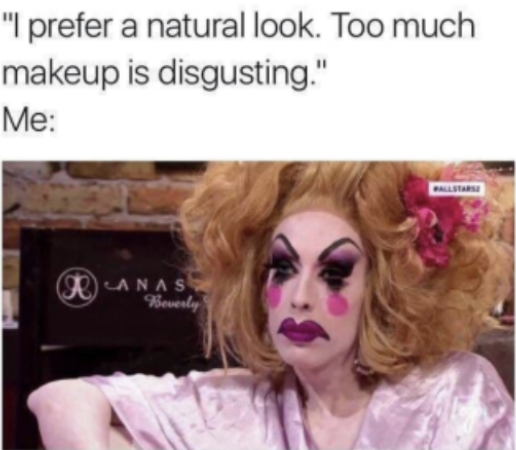}
    \includegraphics[width=0.2\columnwidth]
    {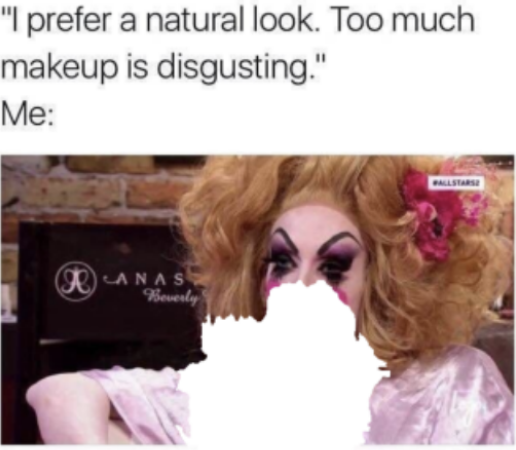}}                          & {\includegraphics[width=0.2\columnwidth]
    {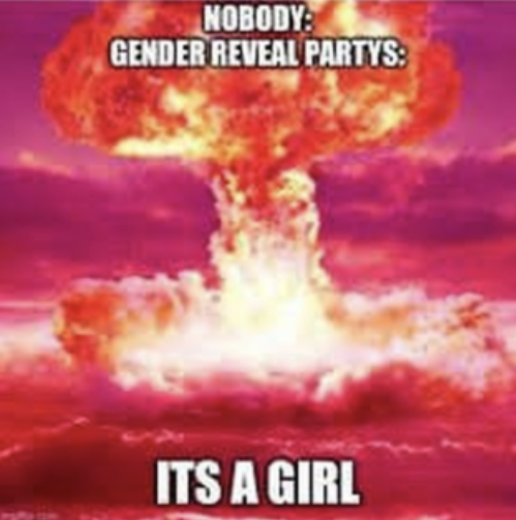}
    \includegraphics[width=0.2\columnwidth]
    {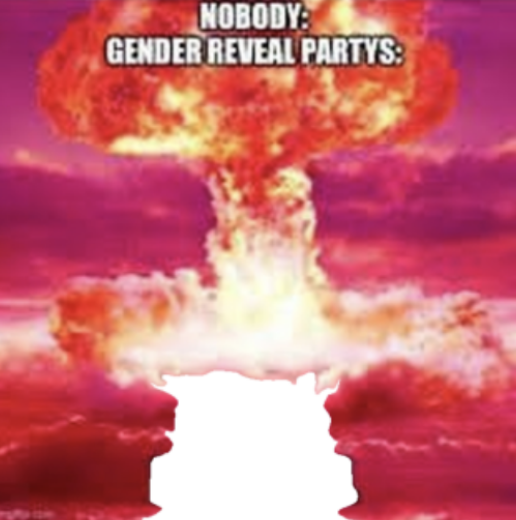}}                            &                                             {\includegraphics[width=0.2\columnwidth]
    {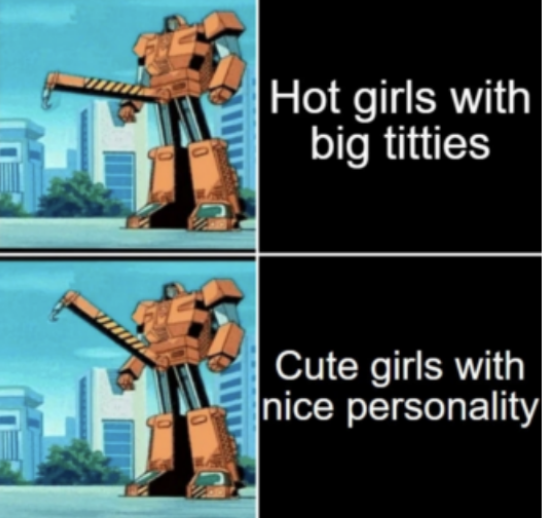}
    \includegraphics[width=0.16\columnwidth]
    {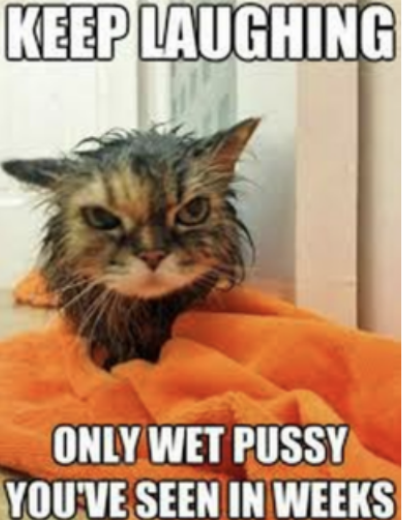}}                 \\ 
                                      &                                            & \multicolumn{2}{c|}{\textbf{CASE 1}}                                                   & \textbf{CASE 2}                                                      \\ \cline{2-5} 
                                      & \multirow{2}{*}{\textbf{Not-Misogynistic}}     & \multicolumn{1}{c}{\includegraphics[width=0.2\columnwidth]
    {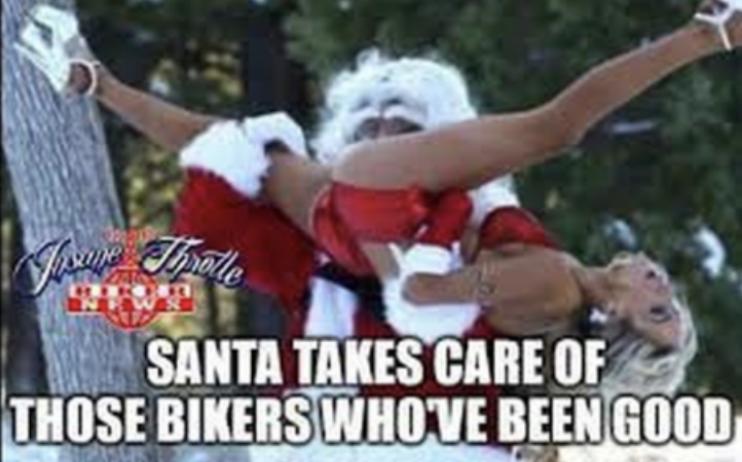}
    \includegraphics[width=0.2\columnwidth]
    {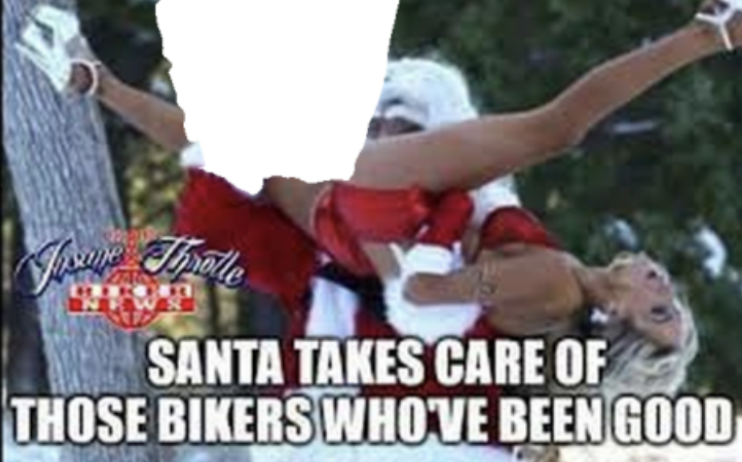}}                           & {\includegraphics[width=0.2\columnwidth]
    {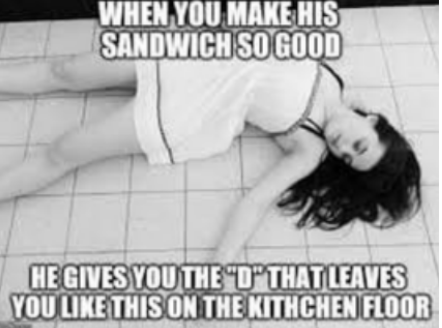}
    \includegraphics[width=0.2\columnwidth]
    {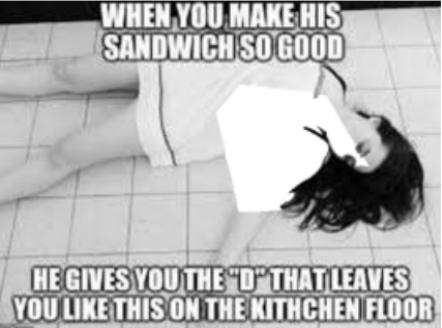}}                            &                                             {\includegraphics[width=0.16\columnwidth]
    {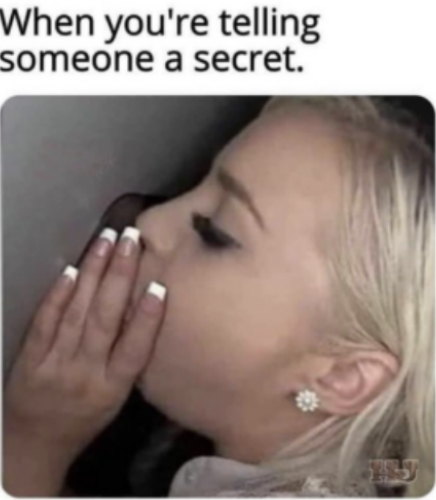}
    \includegraphics[width=0.2\columnwidth]
    {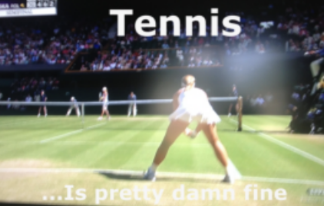}}                 \\ 
                                      &                                            & \multicolumn{2}{c|}{\textbf{CASE 3}}                                                   & \textbf{CASE 4}                                                      \\ \hline
\multirow{4}{*}{\textbf{HARM-C \& P}} & \multirow{2}{*}{\textbf{Harmful}}      & \multicolumn{1}{c}{\includegraphics[width=0.2\columnwidth]
    {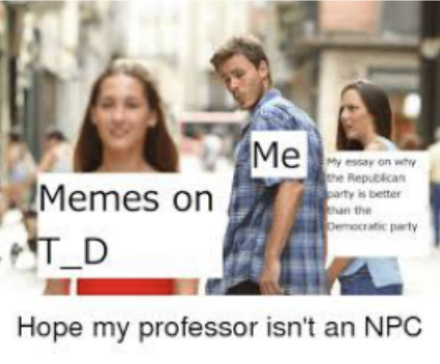}
    \includegraphics[width=0.2\columnwidth]
    {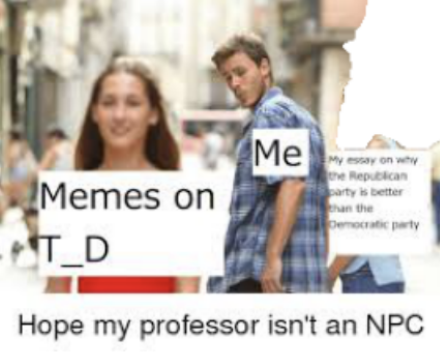}}                           & {\includegraphics[width=0.2\columnwidth]
    {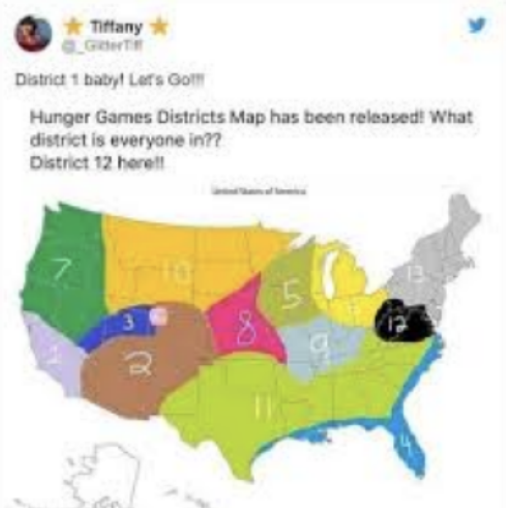}
    \includegraphics[width=0.2\columnwidth]
    {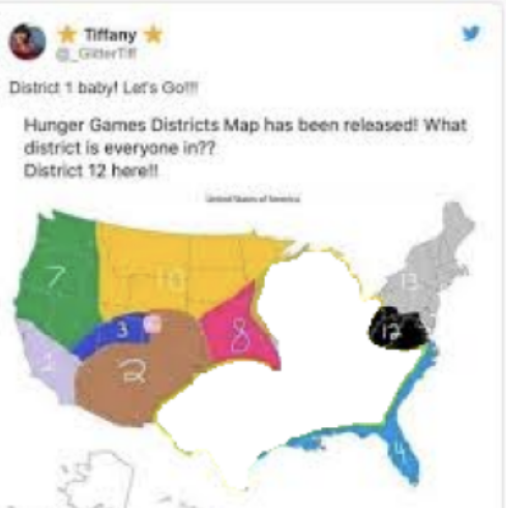}}                           &                                                     {\includegraphics[width=0.18\columnwidth]
    {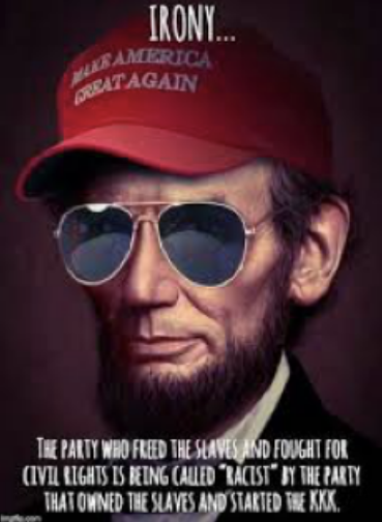}
    \includegraphics[width=0.2\columnwidth]
    {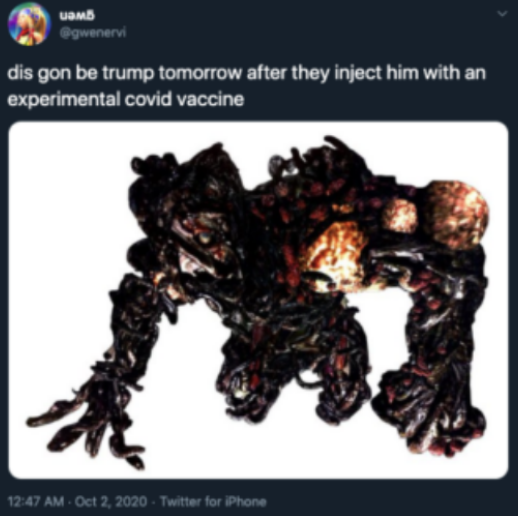}}        \\ 
                                      &                                            & \multicolumn{2}{c|}{\textbf{CASE 1}}                                                   & \textbf{CASE 2}                                                      \\ \cline{2-5} 
                                      & \multirow{2}{*}{\textbf{Not-Harmful}}          & \multicolumn{1}{c}{\includegraphics[width=0.2\columnwidth]
    {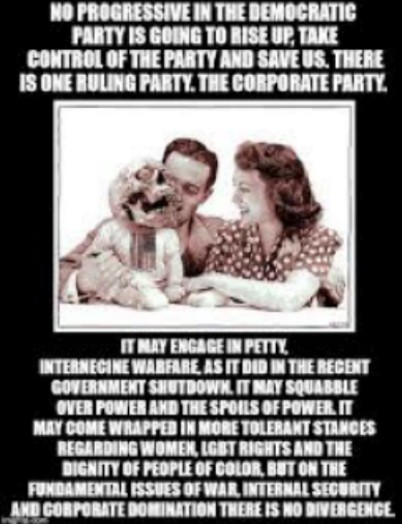}
    \includegraphics[width=0.2\columnwidth]
    {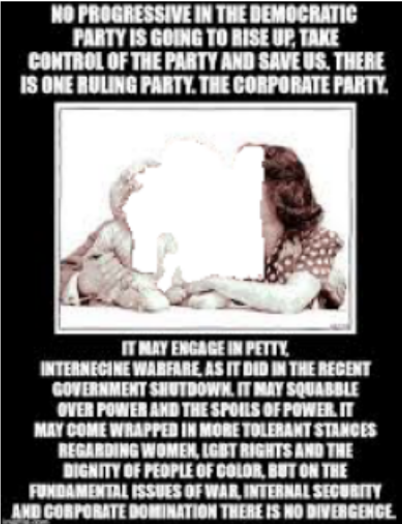}}                           & {\includegraphics[width=0.2\columnwidth]
    {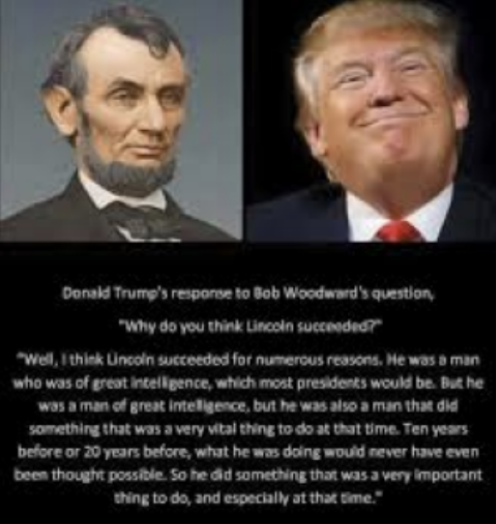}
    \includegraphics[width=0.2\columnwidth]
    {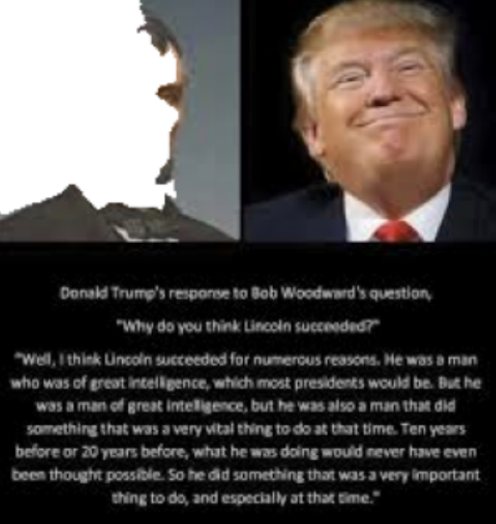}}                            &                                            {
    \includegraphics[width=0.2\columnwidth]
    {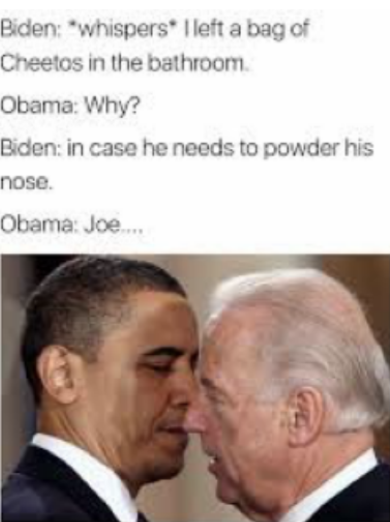}
    \includegraphics[width=0.2\columnwidth]
    {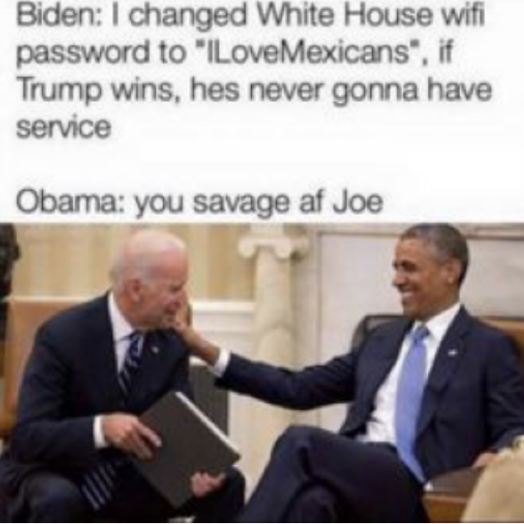}}                  \\ 
                                      &                                            & \multicolumn{2}{c|}{\textbf{CASE 3}}                                                   & \textbf{CASE 4}                                                      \\
\end{tabular}
\caption{\footnotesize\textsc{\textbf{Case Studies}}: Occlusion based predictions. The occlusion is implemented by making a given superpixel white. For each case, we present one sample each from \sysGPT{} \& \sysL{} 13B, respectively. For CASE 1 and CASE 3 we provide both the occluded and non-occluded samples. Due to space constraints and no robust conclusion from manual annotation of BHM and \textsc{HinGlish}, we omit them.}
\label{tab:super_pixel_patch_results}
\end{table*}

\noindent \textbf{CASE 1}: \textit{\textbf{FHM dataset}}: Around one-fourth of the memes are made up of multiple images stacked together. These memes put humans and animals \textit{(apes/gorilla/goat)} in the same frame. Further, certain not hateful memes also seem to contain profane words or mention the common target words either as embedded text or as a representation in the image itself (refer Table~\ref{tab:slic_quantized_result}'s caption for further details). Occlusion results in correct predictions due to the removal of these confusing regions from the meme where the model was misfocusing. That said, our manual inspection indicates that there are memes which are indeed wrongly annotated as not hateful and the predictions of the model for the original meme are arguably correct. \textit{\textbf{MAMI dataset}}: Majority of the memes are indeed wrongly annotated or contain perturbed faces of women with weird makeups or portray men either with (i) women or with (ii) embedded text containing words like \textit{`women', `girlfriend', `girl'.} Moreover, for \sysL{} 13B, some memes are made up of multiple images stacked together. However, the overall theme of the meme is not misogynistic. When occlusion removes the perturbed faces of women or words from the embedded text, the focus of the model is no longer misdirected thus leading to correct predictions. \textit{\textbf{HARM P+C dataset}}: Here again most of the memes are composed of stacked images. Further many of these memes have long text with small font size embedded on them. Such images are even hard for human judges to label. Owing to this very complex nature of the memes, there in no regular pattern indicating why occluding certain parts of the image results in the correct prediction. This is a case where the occlusion based prediction changes are insufficient in explaining the performance gap of the models and more research is needed in the future. \textit{\textbf{BHM +}} \textit{\textsc{{HinGlish}} datasets}: No conclusive pattern was visible due to the diverse topicality of discussion and outdated conversations within current context. However, majority of the memes were composed of stacked images.\\
\noindent \textbf{CASE 2}: \textit{\textbf{FHM dataset}}: Surprisingly, we find that a major portion of the memes are indeed hateful and seem to be incorrectly annotated as not hateful. Common targets include religion, gender, race and politicians. Amongst religion, \textit{`Islam'} is mostly targeted while \textit{`Hitler'} and \textit{`Trump'} are the most targeted politicians. None of the occlusions resulted in a change in the predictions which further reinforces the possibility that the data might be wrongly annotated. \textit{\textbf{MAMI dataset}}:  Majority of the memes pose nudity, vulgarity, feminism amongst other attacks on women. Embedded texts have vulgar words like \textit{`bra', `va**na', `t*ts', `s*xy', `a*s'} targeting women. These memes indeed portray explicit misogyny and as per our analysis, model correctly classifies it as misogynistic and this decision does not get reverted due to occlusion. Here again, we conclude that annotations themselves are incorrect. \textit{\textbf{HARM P+C dataset}}: Here too we manually observe that most of the memes are indeed harmful and are possibly incorrectly annotated. The predictions of the model seem to be correct and occlusions do not change the predictions. \textit{\textbf{BHM +}} \textit{\textsc{{HinGlish}} datasets}: Similar to CASE 1, no conclusive pattern was visible and majority of the memes were composed of stacked images.\\
\textbf{CASE 3}: \textit{\textbf{FHM dataset}}: In this group, most of the memes have very small font size of the embedded text. Further the image has multiple objects or numerous color variations. This confuses the model leading to wrong predictions. Occlusion of these confusing regions allowed the model to focus on the parts of the image important for correct classification. \textit{\textbf{MAMI dataset}}: In most of the cases, image portrays nudity or other forms of vulgarity. In some memes, the embedded text contains the words like \textit{`MILF', `VIRGIN', etc..} targeting women. Occlusion brings the focus of the model to these disturbing elements of the image leading to the correct prediction. \textit{\textbf{HARM P+C dataset}}: Majority of the memes contain the image of \textit{`Trump'} or mention the words \textit{`Trump', `Covid-19}' or \textit{`Corona'}. Length of embedded texts are very large, which possibly confuses the model. Occlusion helps to bring back the focus of the model to the correct regions resulting in correct predictions. \textit{\textbf{BHM +}} \textit{\textsc{{HinGlish}} datasets}: While no conclusion was visible for \textsc{HinGlish}, in BHM majority of the memes were composed of stacked images.\\
\begin{figure*}[!ht]
    \centering
    \includegraphics[width=2.13\columnwidth]{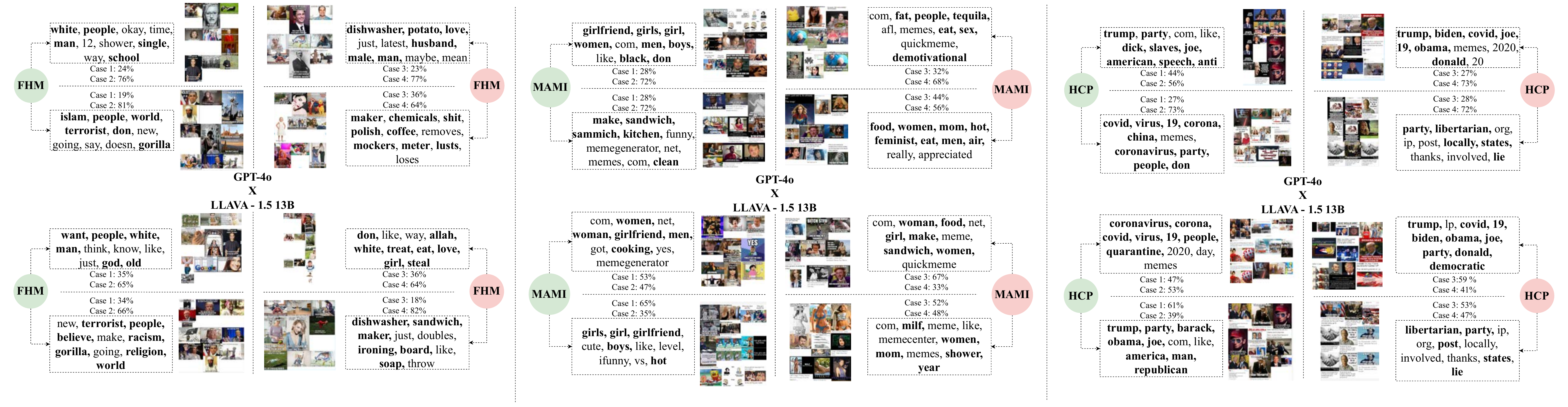}
    \caption{\footnotesize\textsc{\textbf{Typology:}} Green circles represent misclassification to positive label; red signifies misclassification to negative label. Each set of misclassification is bifurcated into two clusters. Distribution of cases, topic words and representative image cluster are shown for \sysGPT{} and \sysL{} 13B models, for FHM, MAMI and HARM-C + P datasets. \textbf{Enlarged image clusters} and results for BHM and \textsc{HinGlish} are in Appendix. Important keywords in each topic are marked in \textbf{bold}.}
    \label{fig:bertopic}
\end{figure*}
\textbf{CASE 4}: \textit{\textbf{FHM dataset}}: Majority of the memes contain implicit hate. Individually neither the image nor the embedded text in the memes portray any hate. Most text have words like \textit{`dishwater', `sandwich maker', `girl', `wive', `girlfriend'.} The images in these memes have cheerful faces of women with no vulgarity. When both the image and text are taken together they portray hate and, quite naturally, the model has difficulty in identifying this implied semantics even when parts of the image are occluded. \textit{\textbf{MAMI dataset}}: Once again these memes seem to bear implicit misogyny. Words like \textit{`dishwater', `sandwich maker', `kitchen'}, and those referring to implicit body shaming appear in the embedded text. The model does not seem to have the requisite reasoning ability to infer the correct class of the meme and occlusion naturally does not come to any help. \textit{\textbf{HARM P+C dataset}}: Drawing conclusion from this case is tricky. However, we observe a common set of memes that portray fake conversations amongst political leaders. These conversations are implicitly harmful and are present in a satirical manner. The model misclassifies both the original and the occluded memes. \textit{\textbf{BHM +}} \textit{\textsc{{HinGlish}} datasets}: Interestingly, we observed many samples wrongly annotated for \textsc{HinGlish} while BHM contained stacked image based memes.

\subsection{Actionable evaluation: Typology of the error cases}
\label{subsection:8.2}
While the previous section allowed us to obtain reasons for misclassification using the occlusion approach, it is largely manual. In this section we present an automatic method to induce the cases we observed earlier. This investigation is motivated by the concept of actionable evaluation through error typology induction introduced in~\cite{font-llitjos-etal-2005-framework,vilar-etal-2006-error}. 
As a first step, for each dataset, we organise the misclassified data points into two groups -- (a) misclassified as positive (hateful/misogynistic/harmful/offensive) and (b) misclassified as negative. Next for the data points for each group of each dataset we first obtain embeddings of the meme image + OCR text using the \textit{\textbf{clip-ViT-B-32}} model. We then run \textbf{multimodal BERTopic}\footnote{\protect\url{https://maartengr.github.io/BERTopic/getting_started/multimodal/multimodal.html}} on each group and bifurcate them into two clusters. Two clusters are specifically chosen to analyze the distribution of two cases (CASE 1/CASE 2 or CASE 3/CASE 4) which we covered in the previous subsection. We present our results in Figure~\ref{fig:bertopic}\footnote{A similar analysis of the model generated explanations are presented in the Appendix.} and in the rest of the section we attempt to make inferences from the obtained results. Note that here we present the results of English datasets; error typology of BHM and \textsc{HinGlish} is updated in Appendix.\\
\noindent\textbf{(i)} We observe that across each cluster (with corresponding topic words), outputs of \sysGPT{} fall into CASE 2 and CASE 4, which naturally follows from the rigidity analysis we have performed. We observe similar patterns for \sysL{} 13B version for the FHM dataset; for the other datasets, the outputs of \sysL{} 13B fall mostly into CASE 1 and CASE 3. This corroborates with our conclusion from human evaluation done earlier.\\ \textbf{(ii)} The topic words and the corresponding clusters for misclassification to hateful memes for FHM are nearly same for both models. However, for \sysGPT{} the topics obtained are more fine-grained compared to \sysL{} 13B in the following cases -- misclassification to not-hateful class for FHM dataset and misclassification to either of the classes for the other datasets. \\ \textbf{(iii)} For HARM-C + P, we observe that for keywords related to Covid-19, both models tend to misclassify to harmful. Note that this is an important observation, since we arrived at no such conclusion from the human annotations.\\ \textbf{(iv)} The implicit keywords like \textit{dishwater}, \textit{sandwich}, \textit{polish}, and \textit{maker} are instances where current VLMs fail.\\ \textbf{(v)} \sysGPT{} in case of misclassification to not-misogynistic for the MAMI dataset fails to understand the implicit hateful references to words like \textit{food}, \textit{hot}, and \textit{fat}.\\ \textbf{(vi)} Both \sysL{} 13B and \sysGPT{} in case of misclassification to hateful memes topics contain keywords related to religion. Thus non-hateful memes containing common target words can often get misclassified.

\subsection{Action items}
\noindent Overall we believe the above two subsections together provide invaluable insights into what are the systematic error patterns that VLMs are vulnerable to. We summarize our analysis and discuss some action items:\\
\textbf{(i)} \textbf{Annotation quality}: Our study reveals that the annotation quality of datasets is a concerning problem. No matter how fancy and sophisticated the models are, the veracity of the results will remain questionable if one cannot ensure the robustness of annotation quality in future. Systematic phase-wise incremental pilots should be undertaken (similar to~\cite{mathew2020hatexplain}) to improve the annotation quality. At the end of each pilot, the quality of each annotation needs to be reviewed. One particularly important trend of good quality annotation is that the inter-annotator agreement should improve at the end of each pilot. Errors committed by the annotators in the early stages of the pipeline should be pointed to them so that these are not repeated in future. Suitable explanations need to be presented to the annotators so that they can understand their errors better. Continuously underperforming annotators, especially in the early pilot stages need to be removed from the pool. Finally, benchmark tasks can be evaluated against the annotated data after every pilot stage and performance can be monitored. If there is a drop in the performance, the annotations need to be revisited and corrective steps need to be taken.\\
\textbf{(ii)} \textbf{Occlusion as blackbox model audit:} Since social media memes include both images and embedded text, typical interpretability schemes like LIME~\cite{10.1145/2939672.2939778}, SHAP~\cite{inproceedings}, GRADCAM~\cite{8237336} etc. are not quite apt for them. Moreover, these methods cannot be applied on closed source models like \sysGPT{}. In contrast, our simple yet effective occlusion based strategy is able to reveal highly relevant interpretations of why VLMs go wrong and this approach is easily generalizable to other similar downstream tasks.\\
\textbf{(iii)} \textbf{Error typology as safety guard rail:} On social media platforms, the content of memes change frequently. Thus costly finetuning of the models is repeatedly required to adapt to these new changes. The meta-level categories of memes like the topic clusters that we induced are far less volatile and any newly formed hate meme can be easily mapped to one of these clusters. Therefore it should suffice to feed these clusters into the design pipeline of the safety guardrails for the VLMs thus eliminating the need for regular finetuning of these models. Further, it can be brewed to design functionality based trees and preference datasets for VLM alignment. For instance, one can attempt to prepare error type aware preference datasets for an RLHF style safety fine-tuning.
%%%%%%%%%%%%%%%%%%%%%%%%%%%%%%%%%%%%%%%%%%%%%%%%%%%%%%%%%%%%%%%%%%%%%%%%
\section{Conclusion}
We present a comprehensive study of popular VLMs on hateful memes, spanning eight different prompt variants. For this study, we utilize six datasets spanning three languages and covering various toxicity dimensions and observe that model performance varies based on datasets, language and prompts used. Furthermore, we also propose an approach to select the best model and prompt combination that generalizes well over considered datasets \& languages. Finally we present a systematic method to induce a typology of the errors committed by such VLMs which could have a long-term impact on how safeguarding approaches should be built in future.
%%%%%%%%%%%%%%%%%%%%%%%%%%%%%%%%%%%%%%%%%%%%%%%%%%%%%%%%%%%%%%%%%%%%%%%%

\section{Limitations}
Our work has a few limitations. First, although we experimented with various prompt settings to identify misclassification patterns, these prompt variants are not exhaustive, and numerous other variants could be explored. Despite this, we are confident that our range of prompts can unveil the actual performance of VLMs in hate meme detection as they cover various broad meta-aspects. Second, we did not use hate meme datasets tailored explicitly for this task by fine-tuning the VLMs because of the issues pointed out in the introduction section. Finally, all our experiments are performed by manually tuning the temperature parameter which is fixed for a specific model; to perform fair evaluation across datasets and prompt variants. Other values can also be tested. In future, we plan to address these limitations.

%%%%%%%%%%%%%%%%%%%%%%%%%%%%%%%%%%%%%%%%%%%%%%%%%%%%%%%%%%%%%%%%%%%%%%%%

\bibliography{aaai25}

%%%%%%%%%%%%%%%%%%%%%%%%%%%%%%%%%%%%%%%%%%%%%%%%%%%%%%%%%%%%%%%%%%%%%%%%

\section{Ethics statement}
Our analysis refrains from attempting to trace users involved in disseminating hate, and we do not intend to harm any individuals or target communities. All experiments were thoroughly conducted using datasets crafted from prior research. Our primary focus was to assess the efficacy of large VLMs in hate meme detection, aiming to pinpoint potential areas for future enhancement.

%%%%%%%%%%%%%%%%%%%%%%%%%%%%%%%%%%%%%%%%%%%%%%%%%%%%%%%%%%%%%%%%%%%%%%%%

\appendix

\section{Definitions}

\label{sec:dataset_definitions}
The definitions provided below are picked from the corresponding dataset papers.

\subsection{FHM dataset}

\begin{itemize}
    \item \textbf{hateful}: A direct or indirect attack on people based on characteristics, including ethnicity, race, nationality, immigration status, religion, caste, sex, gender identity, sexual orientation, and disability or disease. Attack is defined as violent or dehumanizing (comparing people to non-human things, e.g., animals) speech, statements of inferiority, and calls for exclusion or segregation. Mocking hate crime is also considered hateful.
    \item \textbf{not-hateful}: A meme which is not hateful and follows social norms.
\end{itemize}

\subsection{MAMI dataset}

\begin{itemize}
    \item \textbf{misogynistic}: A meme is misogynous if it conceptually describes an offensive, sexist or hateful scene (weak or strong, implicitly or explicitly) having as target a woman or a group of women. Misogyny can be expressed in the form of shaming, stereotype, objectification and/or violence.
    \item \textbf{not-misogynistic}: A meme that does not express any form of hate against women.
\end{itemize}

\subsection{HARM-C and HARM-P datasets}

\begin{itemize}
    \item \textbf{harmful}: Multi-modal units consisting of an image and a piece of text embedded that has the potential to cause harm to an individual, an organization, a community, or civil society more generally. Here, harm includes mental abuse, defamation, psycho-physiological injury, proprietary damage, emotional disturbance, and compensated public image.
    \item \textbf{not-harmful}: Multi-modal units consisting of an image and a piece of text embedded that does not cause any harm to an individual, an organization, a community, or society more generally.
\end{itemize}

\subsection{BHM dataset}
\begin{itemize}
    \item \textbf{hateful}: If it explicitly intends to denigrate, vilify, harm, mock, abuse any entity based on their gender, race, ideology, belief, social, political, geographical and organizational status.
    \item \textbf{not-hateful}: If it is not hateful and follows social norms. community, or the society more generally.
\end{itemize}

\subsection{\textsc{HinGlish} dataset}
\begin{itemize}
    \item \textbf{offensive}: A meme will be categorized as offensive if it either explicitly or implicitly dehumanizes, degrades, insults, or attacks any individual or group based on attributes, such as gender, nationality, sexual orientation, ethnicity, race, skin color, health condition.
    \item \textbf{not-offensive}: A meme that is not offensive and follows social norms.
\end{itemize}

%%%%%%%%%%%%%%%%%%%%%%%%%%%%%%%%%%%%%%%%%%%%%%%%%%%%%%%%%%%%%%%%%%%%%%%%

\section{Prompt strategies}
\label{sec:appendixA}
We provide a detailed list of templates for the corresponding prompt variants in Table \ref{prompt_variations}.
\begin{table*}[!h]
\scriptsize
\centering
\setlength{\tabcolsep}{1mm}
\begin{tabular}{c|l}

\textbf{Prompt variants}                                                                      & \textbf{Prompt templates}                                                                                                                                                                                                                                                                                                                                                                                                                                                                                                                                                                                                                                                                                                                                                                                                                                                                                                                                                                 \\ \hline
\begin{tabular}[c]{@{}c@{}}Vanilla (input)\\ +\\ Vanilla (output)\end{tabular}               & \begin{tabular}[c]{@{}l@{}}Classify the input meme as `positive\_label' or `negative\_label'. Provide the answer as either `positive\_label' or `negative\_label' only.\\ {\textbf{Example output for `positive\_label' meme : `positive\_label'}}\\ {\textbf{Example output for `negative\_label' meme : `negative\_label'}}\end{tabular}                                                                                                                                                                                                                                                                                                                                                                                                                                                                                                                                                                                                                                                                     \\ \hline
\begin{tabular}[c]{@{}c@{}}Definition (input)\\ +\\ Vanilla (output)\end{tabular}            & \begin{tabular}[c]{@{}l@{}}Consider the following {\textbf{definitions.}}\\ 1. `positive\_label' - {\textbf{``Definition of `positive\_label' corresponding to dataset"}}\\ 2. `negative\_label' - {\textbf{``Definition of `negative\_label' corresponding to dataset"}}\\ Classify the input meme as `positive\_label' or `negative\_label' based on the {\textbf{above definitions}} considering the image.\\ Provide the answer as either `positive\_label' or `negative\_label' only.\\ Example output for `positive\_label' meme : `positive\_label'\\ Example output for `negative\_label' meme : `negative\_label'\end{tabular}                                                                                                                                                                                                                                                                                                                                                                                                \\ \hline
\begin{tabular}[c]{@{}c@{}}OCR (input)\\ +\\ Vanilla (output)\end{tabular}                   & \begin{tabular}[c]{@{}l@{}}Classify the input meme as `positive\_label' or `negative\_label' considering the image as well as the {\textbf{extracted}}\\ {\textbf{text from the image which is delimited by three backticks.}}\\ {\textbf{Extracted text from the image: ```OCR extracted text goes here'''}}\\ Provide your answer in the format: `positive\_label' or `negative\_label'.\\ Example output for `positive\_label' meme : `positive\_label'\\ Example output for `negative\_label' meme : `negative\_label'\end{tabular}                                                                                                                                                                                                                                                                                                                                                                                                                                                                                   \\ \hline
\begin{tabular}[c]{@{}c@{}}OCR \& Definition (input)\\ +\\ Vanilla (output)\end{tabular}     & \begin{tabular}[c]{@{}l@{}}Consider the following {\textbf{definitions}}.\\ 1. `positive\_label' - {\textbf{``Definition of `positive\_label' corresponding to dataset"}}\\ 2. `negative\_label' - {\textbf{``Definition of `negative\_label' corresponding to dataset"}}\\ Classify the input meme as `positive\_label' or `negative\_label' based on the {\textbf{above definitions considering the image}} \\ {\textbf{as well as the extracted text from the image which is delimited by three backticks.}}\\{\textbf{Extracted text from the image: ```OCR extracted text goes here'''}}\\ Provide the answer as either `positive\_label' or `negative\_label' only.\\ Example output for `positive\_label' meme : `positive\_label'\\ Example output for `negative\_label' meme : `negative\_label'\end{tabular}                                                                                                                                                                                                                                                                        \\ \hline
\begin{tabular}[c]{@{}c@{}}Vanilla (input)\\ +\\ Explanation (output)\end{tabular}           & \begin{tabular}[c]{@{}l@{}}Classify the input meme as `positive\_label' or `negative\_label'. Provide the answer as either `positive\_label' or `negative\_label' only\\ with an {\textbf{explanation within 30 words explaining your classification.}}\\ Example output for `positive\_label' meme : `positive\_label' - {\textbf{Explain within 30 words that why you classified this meme as `positive\_label'.}}\\ Example output for `negative\_label' meme : `negative\_label' - {\textbf{Explain within 30 words that why you classified this meme as `negative\_label'.}}\end{tabular}                                                                                                                                                                                                                                                                                                                                                                                                                            \\ \hline
\begin{tabular}[c]{@{}c@{}}Definition (input)\\ +\\ Explanation (output)\end{tabular}        & \begin{tabular}[c]{@{}l@{}}Consider the following {\textbf{definitions}}.\\ 1. `positive\_label' - {\textbf{``Definition of `positive\_label' corresponding to dataset"}}\\ 2. `negative\_label' - {\textbf{``Definition of `negative\_label' corresponding to dataset"}}\\ Classify the input meme as `positive\_label' or `negative\_label' based on the {\textbf{above definitions}} considering the image. Provide your answer \\ as either `positive\_label' or `negative\_label' only with an {\textbf{explanation within 30 words explaining your classification.}}\\ Example output for `positive\_label' meme : `positive\_label' - {\textbf{Explain within 30 words that why you classified this meme as `positive\_label'.}}\\ Example output for `negative\_label' meme : `negative\_label' - {\textbf{Explain within 30 words that why you classified this meme as `negative\_label'.}}\end{tabular}                                                                                                                                                       \\ \hline
\begin{tabular}[c]{@{}c@{}}OCR (input)\\ +\\ Explanation (output)\end{tabular}               & \begin{tabular}[c]{@{}l@{}}Classify the input meme as `positive\_label' or `negative\_label' considering the image as well as the {\textbf{extracted text from the image}} \\ {\textbf{which is delimited by three backticks.}}\\ {\textbf{Extracted text from the image: ```OCR extracted text goes here'''}}\\ Provide your answer in the format: `positive\_label' or `negative\_label', followed by an {\textbf{explanation within 30 words explaining your classification.}}\\ Example output for `positive\_label' meme : `positive\_label' - {\textbf{Explain within 30 words that why you classified this meme as `positive\_label'.}}\\ Example output for `negative\_label' meme : `negative\_label' - {\textbf{Explain within 30 words that why you classified this meme as `negative\_label'.}}\end{tabular}                                                                                                                                                                                                                                              \\ \hline
\begin{tabular}[c]{@{}c@{}}OCR \& Definition (input)\\ +\\ Explanation (output)\end{tabular} & \begin{tabular}[c]{@{}l@{}}Consider the following {\textbf{definitions.}}\\ 1. `positive\_label' - {\textbf{``Definition of `positive\_label' corresponding to dataset"}}\\ 2. `negative\_label' - {\textbf{``Definition of `negative\_label' corresponding to dataset"}}\\ Classify the input meme as `positive\_label' or `negative\_label' based on the {\textbf{above definitions considering the image as well as the extracted}} \\ {\textbf{text from the image which is delimited by three backticks.}}\\ {\textbf{Extracted text from the image: ```OCR extracted text goes here'''}}\\ Provide your answer in the format: `positive\_label' or `negative\_label', followed by {\textbf{an explanation within 30 words explaining your classification.}}\\ Example output for `positive\_label' meme : `positive\_label' - {\textbf{Explain within 30 words that why you classified this meme as `positive\_label'.}}\\ Example output for `negative\_label' meme : `negative\_label' - {\textbf{Explain within 30 words that why you classified this meme as `negative\_label'.}}\end{tabular} \\ \hline
\end{tabular}
\caption{\textbf{\textsc{Employed Prompts:}} Representative examples for different prompt variants. Definition for corresponding labels can be picked from Appendix\ref{sec:dataset_definitions}. `positive\_label' and `negative\_label' will be replaced by corresponding labels as per the dataset. \textit{Note:} Important text in each prompt variant has been indicated in {\textbf{boldface}}.}
\label{prompt_variations}
\end{table*}

%%%%%%%%%%%%%%%%%%%%%%%%%%%%%%%%%%%%%%%%%%%%%%%%%%%%%%%%%%%%%%%%%%%%%%%%

\section{Reproducibility steps}
\label{sec:reproducibility}
We briefly summarize our methodology so that our research can be easily reproduced by the research community:\\
\textbf{Datasets:} All six datasets that we have used are commonly used for \textit{hateful/misogynistic/harmful/offensive} meme detection tasks. The links to these datasets can be found here -- (\textit{FHM})\footnote{\url{https://www.kaggle.com/datasets/parthplc/facebook-hateful-meme-dataset}}, (\textit{MAMI})\footnote{\url{https://github.com/TIBHannover/multimodal-misogyny-detection-mami-2022?tab=readme-ov-file}}, (\textit{HARM-C \& HARM-P})\footnote{\url{https://github.com/LCS2-IIITD/MOMENTA}}, (\textit{BHM})\footnote{\url{https://github.com/eftekhar-hossain/Bengali-Hateful-Memes/blob/main/README.md}}, and (\textsc{HinGlish})\footnote{\url{https://github.com/Gitanjali1801/CM_MEMES}}.\\
\textbf{Processors}: We used the respective model processors to process our images and text. From \textbf{HuggingFace}, we used the \texttt{AutoProcessor.from\_pretrained\\(model\_checkpoint)} API and passed the image and text to the processor before feeding it to the model. Here we passed \texttt{model\_checkpoint} corresponding to the models we used (as discussed in main content).\\
\textbf{Quantization:} We used \textbf{HuggingFace} APIs for quantization\footnote{\url{https://huggingface.co/docs/transformers/en/main_classes/quantization}}. We specifically used \textit{bitsandbytes}\footnote{\url{https://huggingface.co/docs/bitsandbytes/index}} library to setup 8-bit quantization.\\
\textbf{Embedded text (OCR)}: We used the OCR text already present in the dataset for our experiments on memes. For BHM dataset, we translate Bengali OCR text to English using AI4Bharat\footnote{\url{https://ai4bharat.iitm.ac.in/}} models.\\
\textbf{\textsc{BerTopic:}} We set \texttt{random\_state} equal to 42 in \textsc{Umap} while generating clusters.\\
\textbf{License agreement}: We agreed to every license before accessing any model/dataset. Since we used open-source models using HuggingFace, we agreed to their license and for the datasets, we thoroughly read the license and used them as per the corresponding agreement. Please navigate to corresponding mentioned dataset URLs/model checkpoints to get complete license details.

%%%%%%%%%%%%%%%%%%%%%%%%%%%%%%%%%%%%%%%%%%%%%%%%%%%%%%%%%%%%%%%%%%%%%%%%

\section{Examples of ambiguous outputs}
We report four representative example outputs obtained from the VLMs where it was not able to classify amongst \texttt{list\_of\_labels}. Corresponding model, prompt variant and dataset is mentioned for each of these examples. Important text leading to ambiguity is \textbf{highlighted}. Due to these ambiguous outputs, we set a threshold of 90\% support while presenting and comparing our results.\\
\label{sec:outputs_extended}

\noindent{\textbf{(i) \sysI{}: Definition Input + Explanation output (\textit{MAMI} dataset)\\}}
\texttt{{\textbf{Misogynistic}} - The meme depicts a woman running from a car, which is a stereotype that women are weak and need to be protected. The woman is also shown with a large belly, which is an objectification of women's bodies. The meme is sexist and hateful towards women.\\ {\textbf{Not-misogynistic}} - The meme is not misogynistic because it does not depict any form of hate towards women. It is a simple joke that does not have any negative connotations towards women.}

In the above output, model instead of choosing only one label out of \textit{\{Misogynistic or Not-misogynistic\}} and explaining the reason behind its classification, provided an output with both labels. It further went on to justify  both of the labels.\\

\noindent{\textbf{(ii) \sysB{} Vicuna 7B: Vanilla Input + Explanation output (\textit{FHM} dataset)\\}}
\texttt{{\textbf{Example output for hateful meme: hateful}} - This meme is hateful because it makes fun of a woman's appearance and implies that she is not capable of being a good mother.\\ {\textbf{Example output for not-hateful meme: not-hateful}} - This meme is not hateful because it is a lighthearted comment about a woman's appearance and does not imply any negative judgments about her abilities as a mother.}

In the above output, model repeated the example output format provided to it for both labels \textit{\{Hateful and Not-hateful\}}. It then further provided an explanation for both labels corresponding to the input meme.\\

\noindent\textbf{{(iii) \sysB{} Vicuna 7B: OCR Input + Vanilla output \textit{(FHM Dataset)}\\}}
\texttt{The image contains a man who is eating a banana, and the text "bravery at its finest" is written above it. Based on the context and the extracted text, it is not possible to determine whether the meme is \textbf{hateful or not-hateful}}.

In the above output, the model does not adhere to the desired output format. Instead of generating only the label out of \textit{\{Hateful or Not-hateful\}}, it gives a brief explanation and finally infers that it is not possible to determine whether the meme is hateful or not.\\

\noindent\textbf{{(iv) \sysB{} FlanT5-xl: OCR Input + Explanation output (\textit{HARM-C} dataset)\\}}
\texttt{The message from President Trump to Covid-19 is a meme that is a warning to people to leave the country. \textbf{The message is not harmful, but it is not harmless}}.

Here, the model's output does not adhere to the desired output format. Moreover, model diplomatically bypassed the classification task by considering meme to be neither harmful nor harmless.

%%%%%%%%%%%%%%%%%%%%%%%%%%%%%%%%%%%%%%%%%%%%%%%%%%%%%%%%%%%%%%%%%%%%%%%%

\section{Multimodal BERTopic image clusters}
\begin{figure*}[!ht]
    \centering
    \includegraphics[width=2.0\columnwidth]{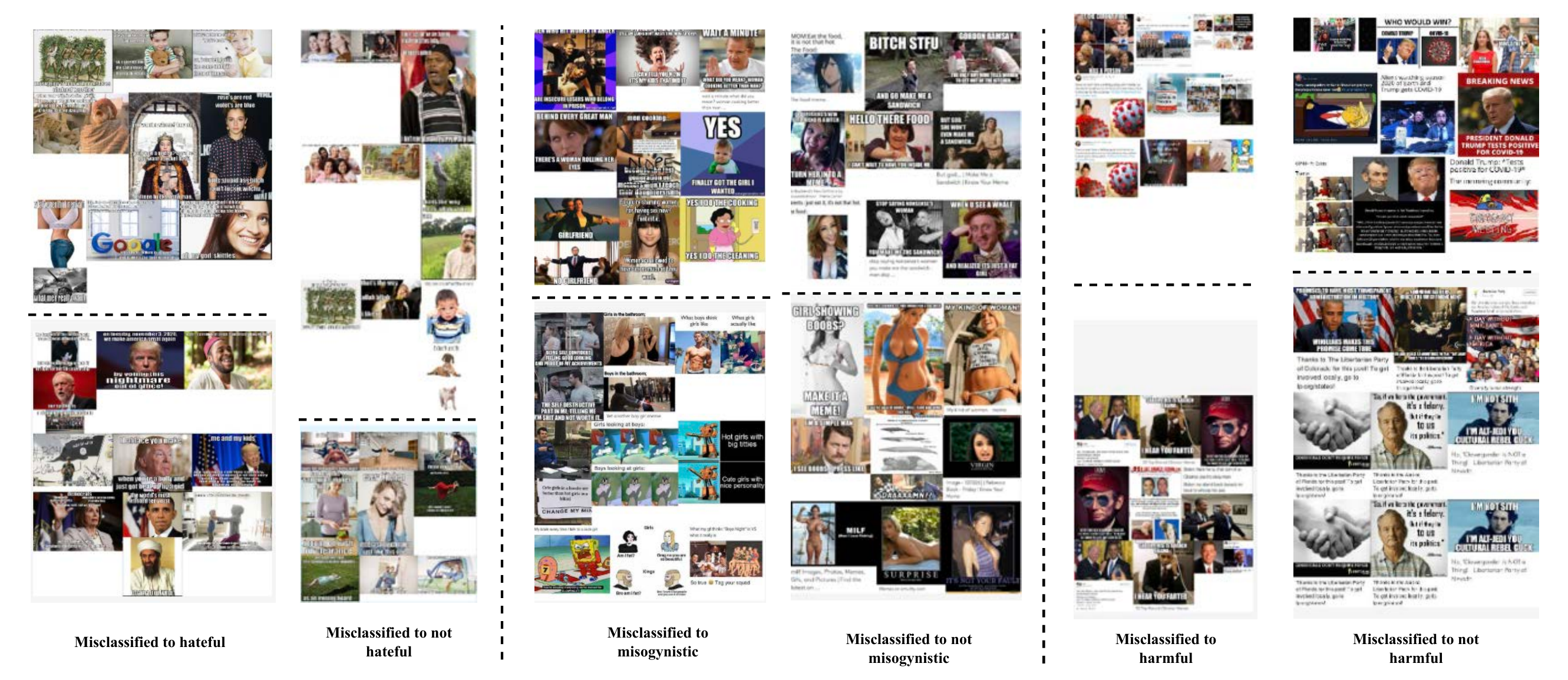}
    \caption{\textbf{\textsc{Typology Clusters:}} Upper panel images: \sysGPT{} error clusters. Lower panel images: \sysL{} 13B error clusters. Each cluster and dataset are separated by dashed lines.}
    \label{fig:bertopic_clusters}
\end{figure*}

The image clusters of the error typology induction for FHM, MAMI and HARM C+P datasets are presented in Figure~\ref{fig:bertopic_clusters}.
\label{sec:image_clusters}
\begin{figure*}[!ht]
    \centering
    \includegraphics[width=2.0\columnwidth]{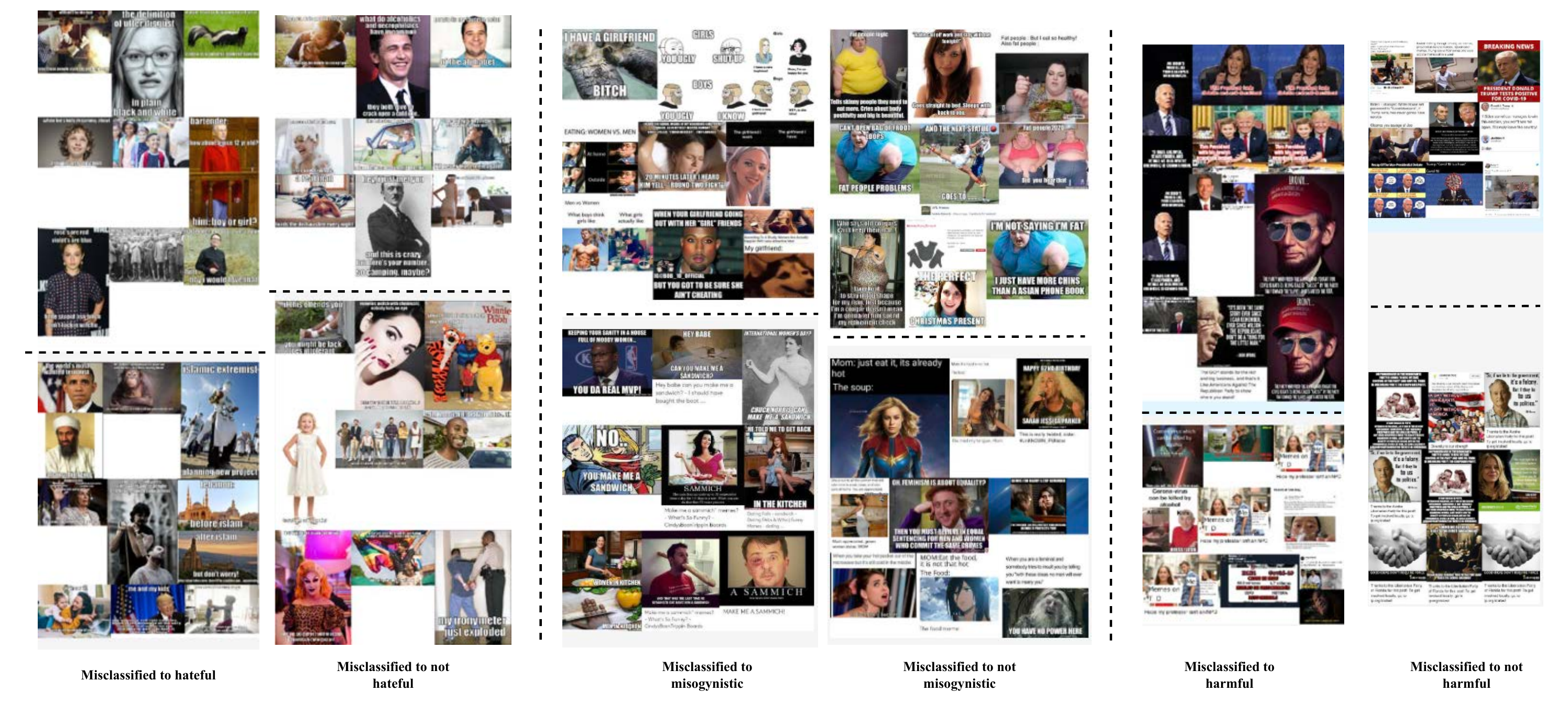}
\end{figure*}

\section{BHM and \textsc{HinGlish} typology}
\begin{figure*}[!ht]
    \centering
    \includegraphics[width=2\columnwidth]{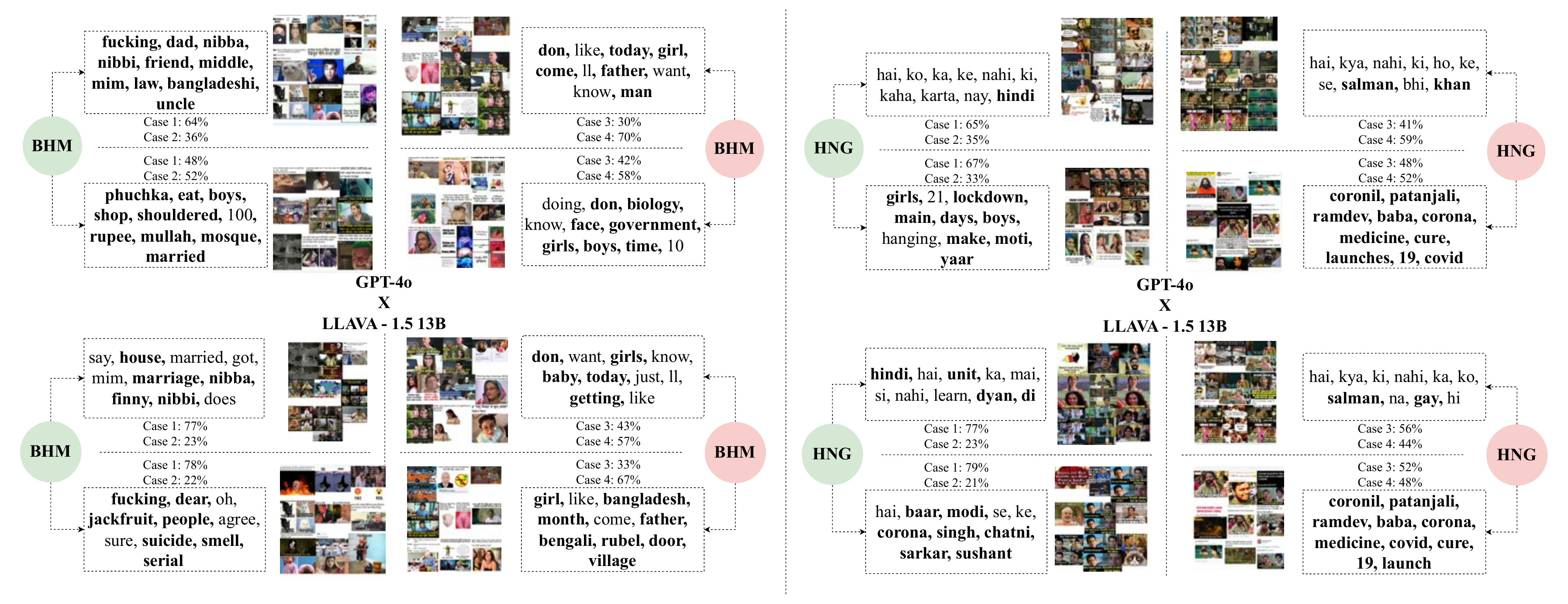}
    \caption{\footnotesize\textsc{\textbf{Typology Extended:}} Green circles represent misclassification to positive label; red signifies misclassification to negative label. Each set of misclassification is bifurcated into two clusters. Distribution of cases, topic words and representative image cluster are shown for \sysGPT{} and \sysL{} 13B models, for BHM and \textsc{HinGlish} datasets. Important keywords in each topic are marked in \textbf{bold}.}
    \label{fig:bertopic_appendix}
\end{figure*}
From Figure~\ref{fig:bertopic_appendix}, we surprisingly observe that both models seemingly correctly classify to offensive for BHM dataset since the induced topic -- \textit{f*cking, mullah, nibba, and nibbi} are indeed used to propagate hateful sentiments and are prevalent within Indian subcontinent. Also, the decreased distribution of CASE 2 and CASE 4 compared to English datasets for \sysGPT{} signifies its weakness for multilingual data. For \textsc{HinGlish}, most clusters contain common pronouns used in \textsc{HinGlish} language. However, for both models, the second cluster corresponding to cases of misclassification to the not offensive class has keywords related to Covid-19; this again corroborates the earlier observation from the occlusion based experiments.

Therefore, the extended study on BHM and \textsc{HinGlish} further demonstrates the importance of our automatic typology induction.

\section{Annotation guidelines}

The annotation process was carried out to evaluate the effect of occlusion of misclassified memes across four cases. For each case, annotators had to bifurcate every meme into systematic error cases. These cases were defined by the authors of the paper after having a broad review of all misclassification cases. Further, it was observed that each case broadly focused on specific classes. Annotators were asked to assess each of the cases alongwith occlusion results to help us better understand the specific regions of the meme important for the models.

\noindent Four separate files were created, each corresponding to one of these cases, to document these instances and understand the nuances of model errors. Each page of the file contains the \textit{image ID}, the \textit{original image}, its \textit{ground truth label}, the \textit{occluded images} that changed the original prediction, and the \textit{corresponding predictions} after occlusion. The evaluation process identified and categorized the misclassified memes into nine distinct error situations based on their visual and textual characteristics:

\begin{enumerate}
    \item \textbf{NHM (Not hateful memes)}: Memes featuring common target words (e.g., ``Islam,'' ``migrants,'' ``white''), political figures, or profane language.
    \item \textbf{WA (Wrong annotation)}: Memes incorrectly labeled in the dataset.
    \item \textbf{SI (Stacked images)}: Memes with overlaid or stacked visual elements.
    \item \textbf{MCVG (Multiple color variations or grayscale)}: Memes containing varied color schemes, grayscale visuals, or multiple objects.
    \item \textbf{URET (Unreadable embedded text)}: Memes with lengthy or small-font embedded text that is difficult to read.
    \item \textbf{IM (Implicit hate meme)}: Memes that convey hate or harmful content implicitly, where the harmful intent is not straightforward to understand.
    \item \textbf{PFW (Perturbed or animated faces of women)}: Memes featuring distorted, exaggerated, or animated female faces.
    \item \textbf{NV (Nudity or vulgarity)}: Memes containing explicitly vulgar/nude visuals or embedded text.
    \item \textbf{FC (Fake conversation)}: Memes designed as fabricated dialogues or text exchanges.
\end{enumerate}

The results of the annotation process, are summarized in the main content of the paper and we have provided detailed insights into the distribution of these situations across the four cases for both \sysGPT{} and\sysL{} 13B models. For instance:
\begin{itemize}
    \item \textbf{CASE 1}: Prominently featured NHM, WA, PFW, and SI, indicating these categories were sensitive to occlusion.
    \item \textbf{CASE 2}: Largely influenced by WA and NHM, indicating wrong annotation in datasets and stereotyping common target words.
    \item \textbf{CASE 3}: Showed issues with NV and MCVG, where models struggled with explicit content or complex visual layouts.
    \item \textbf{CASE 4}: Highlighted errors in processing FC and IM, emphasizing limitations in understanding fabricated contexts or subtle cues.
\end{itemize}

\noindent The annotation process sheds light on the limitations of VLMs when handling nuanced or visually complex memes. By categorizing errors and analyzing the impact of occlusion, the study underscores the importance of improving feature extraction and capturing subtle contextual cues to enhance model robustness and predictive accuracy.

\section{Takeaways from prompt variants}
While we have covered extensive experiment and analysis in the main content, we summarize key observations here as well.\\
\textbf{(i)} Open source models tend to perform much better when vanilla as output is prompted. They seemingly confuse with the generation of explanation.\\
\textbf{(ii)} OCR text alone helps \sysI{} and the incorporation of definition generally degrades its results.\\
\textbf{(iii)} OCR + definition works best for \sysL{} models. Surprisingly, 7B version works best with explanation as output while 13B works best with vanilla as output.\\
\textbf{(iv)} We can safely say that although open source VLMs might be good at generation, their performance as predictors needs improvement. Our study proves that although open source VLMs are good generators, a long path awaits before they match with likes of \sysGPT{}.\\
\textbf{(v)} We also observed that the models which generally perform better across all datasets are also more stable; hence \sysGPT{} had the least standard deviation across all models considered.\\
\textbf{(vi)} On multilingual datasets, even \sysGPT{} struggles in terms of performance.\\
\textbf{(vii)} \sysB{} models generate very large outputs which leads to ambiguity. This behaviour considerably increases in case of explanation as output.

%%%%%%%%%%%%%%%%%%%%%%%%%%%%%%%%%%%%%%%%%%%%%%%%%%%%%%%%%%%%%%%%%%%%%%%%

\section{Examples of wrong annotation}
\begin{figure*}[!ht]
    \centering
    \includegraphics[width=1.65\columnwidth]{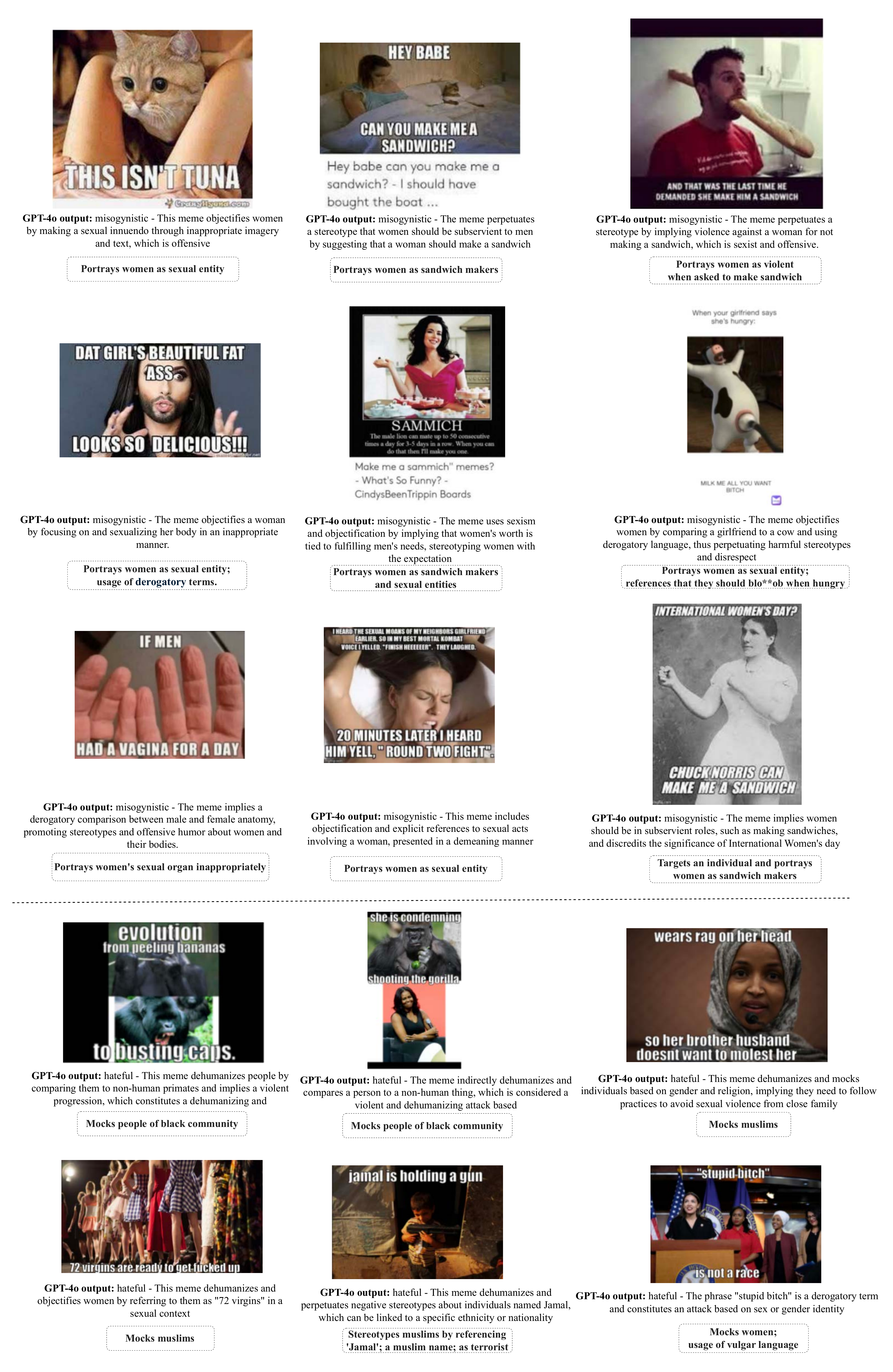}
    \caption{\footnotesize\textsc{\textbf{Examples of Wrong Annotation}}: Fifteen examples, nine from MAMI and six from FHM are shown for CASE 2 of \sysGPT{} with def + OCR as input and explanation as output. Output of the model is also provided.}
    \label{fig:examples}
\end{figure*}

Figure~\ref{fig:examples} presents fifteen memes; nine from MAMI and six from FHM. All these memes are marked as either misogynistic or hateful by human annotators employed in our work. All these fall in CASE 2 i.e., none of the occluded samples were able to correct the prediction. We can observe that \sysGPT{} does not only predict these cases correctly, but also provides reasonable justification.

%%%%%%%%%%%%%%%%%%%%%%%%%%%%%%%%%%%%%%%%%%%%%%%%%%%%%%%%%%%%%%%%%%%%%%%%

\section{Analysis of \sysGPT{} explanations}
\begin{figure*}[!ht]
    \centering
    \includegraphics[width=2\columnwidth]{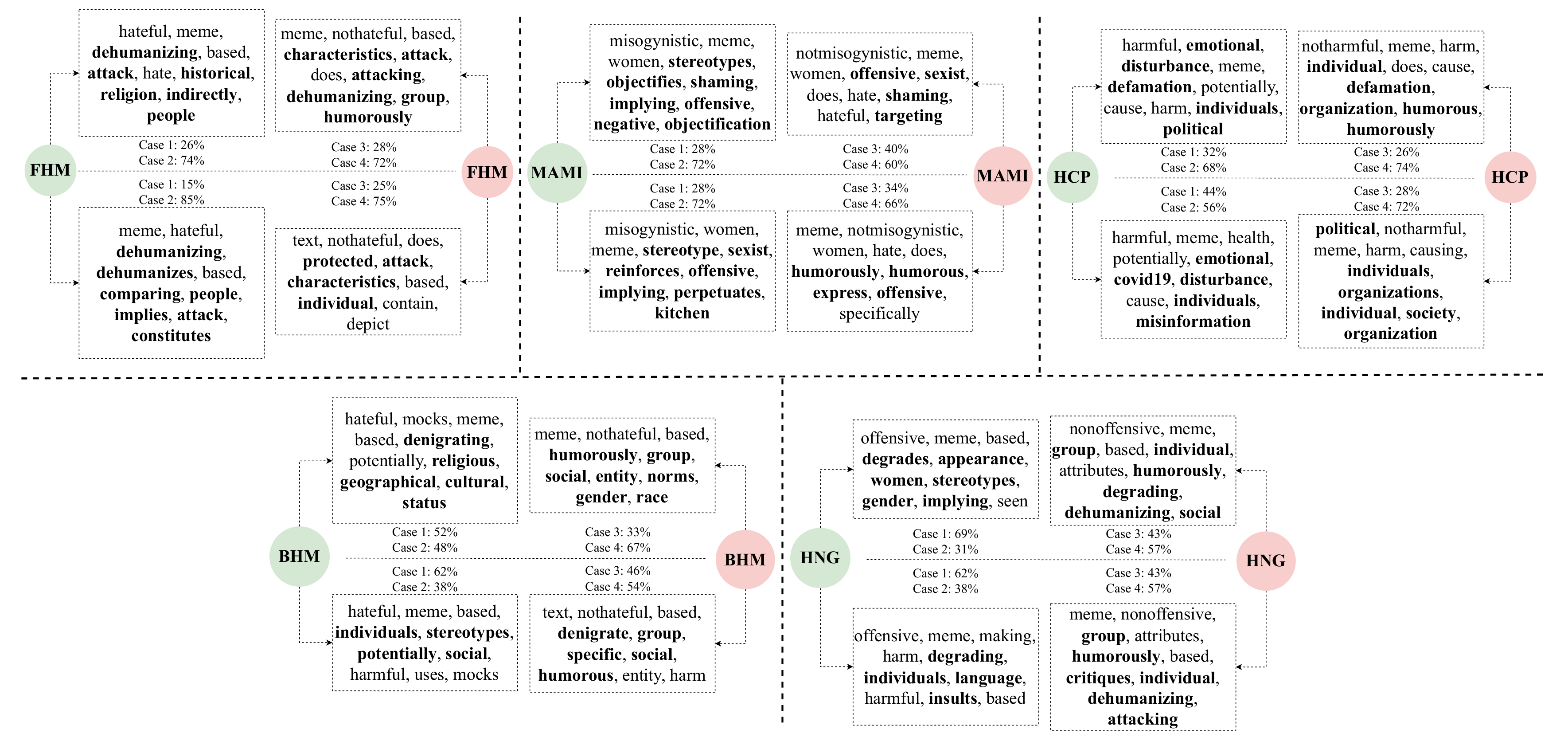}
    \caption{\footnotesize\textsc{\textbf{Typology Explanation:}} Green circles represent misclassification to positive label; red signifies misclassification to negative label. Each set of misclassification is bifurcated into two clusters. Distribution of cases and topic words are shown for \sysGPT{} for FHM, MAMI, HARM-C + P, BHM and \textsc{HinGlish} datasets. Important keywords in each topic are marked in \textbf{bold}.}
    \label{fig:bertopic_explanation}
\end{figure*}
In Figure~\ref{fig:bertopic_explanation}, we provide the induced typology of \sysGPT{} explanations for misclassified samples. We use the default BERTopic pipeline to generate typology alongwith clusters and summarize our key observations:\\
\textbf{(i)} The induced topics are not very diverse compared to what we obtained previously (refer to Figures~\ref{fig:bertopic} and \ref{fig:bertopic_appendix}). Although these topics provide insights into the generation capability of the model, they cannot be effectively brewed as safety guardrails.\\
\textbf{(ii)} From the misclassification to negative labels (red circles in Figure~\ref{fig:bertopic_explanation}), we observe that models generate words like \textit{`humorously'} and \textit{`humorous'}. Drawing from occlusion based analysis we conclude that models generally consider implicitly hateful memes as humorous and are unable to identify hidden hateful elements.\\
\textbf{(iii)} Further we also observe from the misclassification to negative labels that the model is seemingly over aggressive to classify memes targeting \textit{`group', `society'} and \textit{`organization'} as not-(hateful/harmful/offensive).\\
\textbf{(iv)} The topic \textit{`stereotype'} is only present in misclassification to positive label for MAMI, BHM and \textsc{HinGlish} datasets. Interestingly, \textit{`religion'} and \textit{`religious'} also fall under misclassification to positive label for FHM \& BHM (both datasets contain hateful memes). This reinforces the observations from our occlusion based study whereby the models seem to excessively stereotype religion and forcefully misclassify.\\
\textbf{(v)} Another interesting observation is the presence of words like \textit{`implies'} and \textit{`implying'} only in the topic clusters for miscalssification to positive labels -- this means that models like \sysGPT{} attempt to stress on defending its classification to positive labels.\\
\textbf{(vi)} \sysGPT{} when given definition as input generates relevant explanations, containing the words like \textit{`dehumanizing', `objectification', `stereotypes', `shaming'} and \textit{`defamation'} among many others. This tells us that after analyzing the image and OCR text, \sysGPT{} is able to lucidly apply reasoning to generate task specific and relevant output.\\
\textbf{(vii)} Surprisingly, the case distribution of clusters highly align with what we concluded in actionable evaluation study. This concludes that the rigidness distribution is just not limited to inputs but is propagated to outputs as well. 

%%%%%%%%%%%%%%%%%%%%%%%%%%%%%%%%%%%%%%%%%%%%%%%%%%%%%%%%%%%%%%%%%%%%%%%%

\end{document}